%% file: main.tex
\documentclass[10pt,twocolumn,letterpaper]{article}

\usepackage[pagenumbers]{cvpr} %

\usepackage{graphicx}
\usepackage{amsmath}
\usepackage{amssymb}
\usepackage{booktabs}
\usepackage{color}
\usepackage{multirow}
\usepackage{xcolor}
\usepackage{balance}

\newcommand\rot[1]{\rlap{\rotatebox{45}{#1}}}
\newcommand\OK{$\checkmark$}

\usepackage[pagebackref,breaklinks,colorlinks]{hyperref}

\usepackage[capitalize]{cleveref}
\crefname{section}{Sec.}{Secs.}
\Crefname{section}{Section}{Sections}
\Crefname{table}{Table}{Tables}
\crefname{table}{Tab.}{Tabs.}

\definecolor{rred}{RGB}{245, 152, 153}
\definecolor{oorange}{RGB}{253, 205, 154}
\definecolor{yyellow}{RGB}{248,244,140}

\usepackage[capitalize]{cleveref}
\crefname{section}{Sec.}{Secs.}
\Crefname{section}{Section}{Sections}
\Crefname{table}{Table}{Tables}
\crefname{table}{Tab.}{Tabs.}

\DeclareMathOperator*{\argmin}{argmin}

\def\cvprPaperID{454} %
\def\confName{CVPR}
\def\confYear{2023}

\begin{document}

\title{RANA: Relightable Articulated Neural Avatars\vspace{-2mm}}

\author{Umar Iqbal\textsuperscript{1*} \quad 
        Akin Caliskan\textsuperscript{2*} \quad
        Koki Nagano\textsuperscript{1} \quad 
        Sameh Khamis\textsuperscript{1} \quad 
        Pavlo Molchanov\textsuperscript{1} \quad 
        Jan Kautz\textsuperscript{1} \\
\textsuperscript{1}NVIDIA \qquad \textsuperscript{2}University of Surrey, UK \\
\url{https://nvlabs.github.io/RANA/}
}

\newcommand\extrafootertext[1]{%
    \noindent
    \bgroup
    \renewcommand\thefootnote{\fnsymbol{footnote}}%
    \renewcommand\thempfootnote{\fnsymbol{mpfootnote}}%
    \footnotetext[0]{#1}%
    \egroup
}

\input{figures/fig_teaser}
\maketitle
\extrafootertext{\hspace{-5mm}*equal contribution. The work was partially done during AC's internship at NVIDIA.}
\input{sec/0_abstract}
\input{sec/1_introduction}

\input{sec/2_related}

\input{sec/3_method}
\input{sec/4_experiments}

\input{sec/5_conclusion.tex}

\appendix
\input{appendix_content.tex}

{\small
\balance
\bibliographystyle{ieee_fullname}
\bibliography{avatar}
}
\clearpage
\end{document}

%% file: figures/fig_teaser.tex
\twocolumn[{%
\renewcommand\twocolumn[1][]{#1}%
\vspace{-2em}
\maketitle
\thispagestyle{empty}
\vspace{-3em}
\begin{center}
    \centering
\scalebox{0.97}{
\small
    \setlength{\tabcolsep}{1pt}
    \renewcommand{\arraystretch}{0.4}
  \begin{tabular}{ccccc}
    \includegraphics[width=0.2\textwidth]{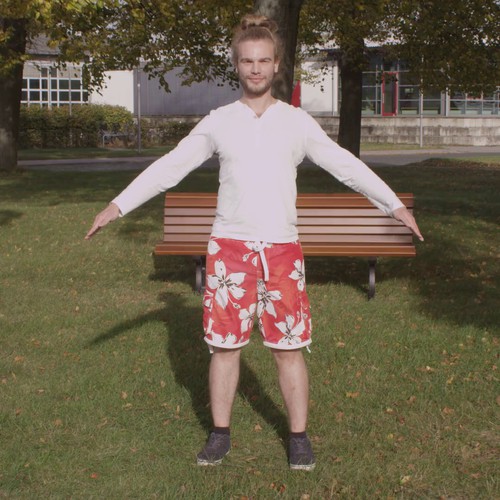}
    \includegraphics[width=0.2\textwidth]{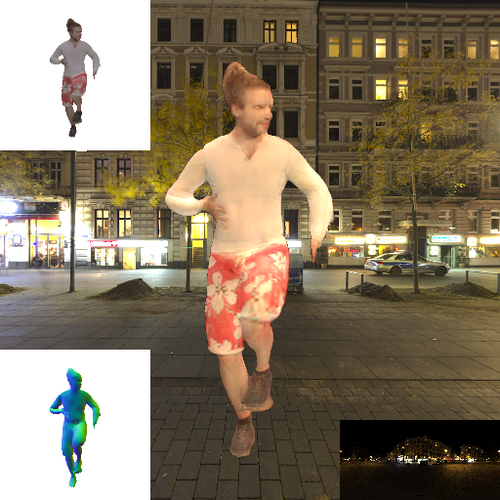} &
    \includegraphics[width=0.2\textwidth]{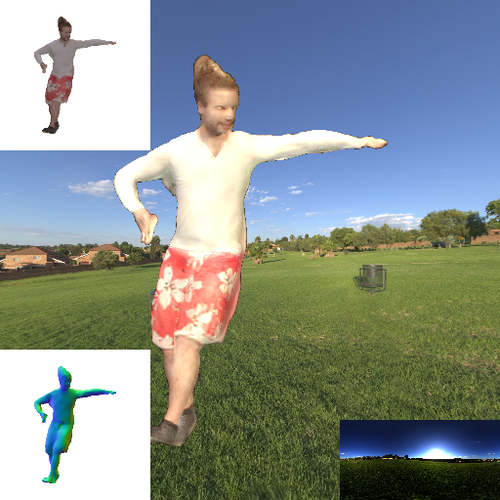} &
    \includegraphics[width=0.2\textwidth]{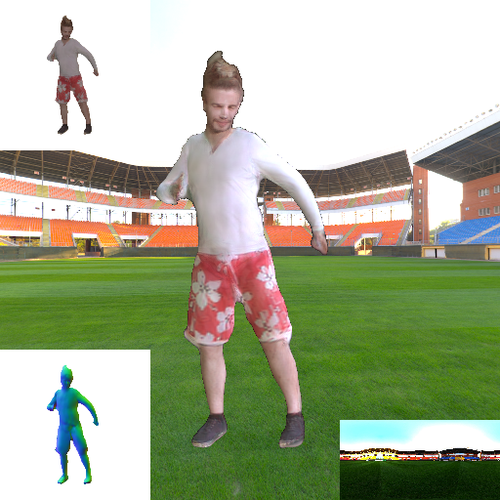} &
    \includegraphics[width=0.2\textwidth]{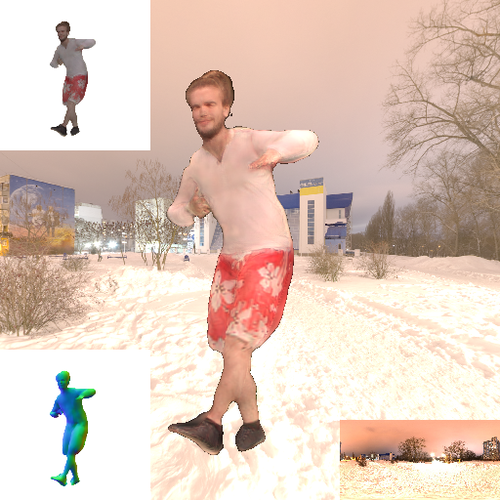} \\
    \includegraphics[width=0.2\textwidth]{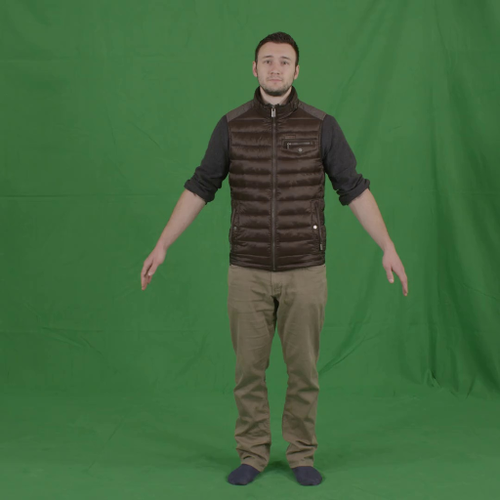} 
    \includegraphics[width=0.2\textwidth]{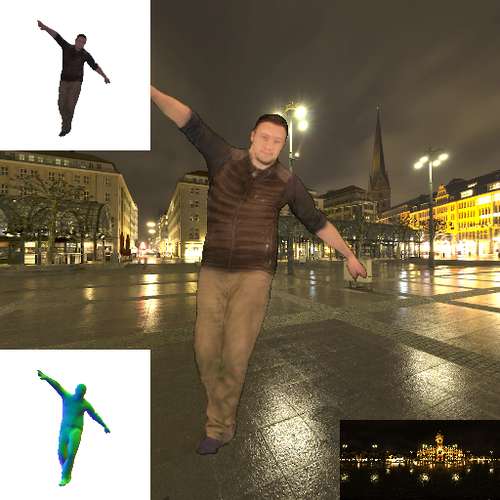} &
    \includegraphics[width=0.2\textwidth]{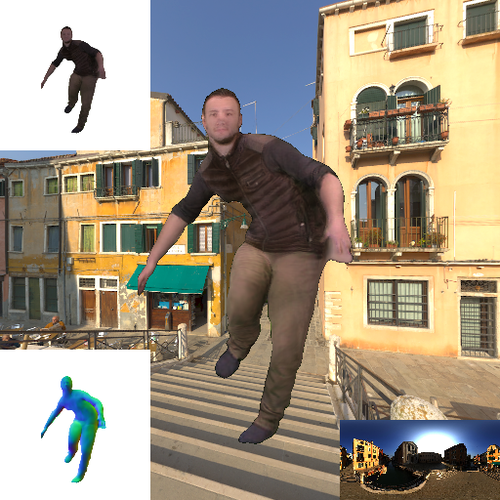} &
    \includegraphics[width=0.2\textwidth]{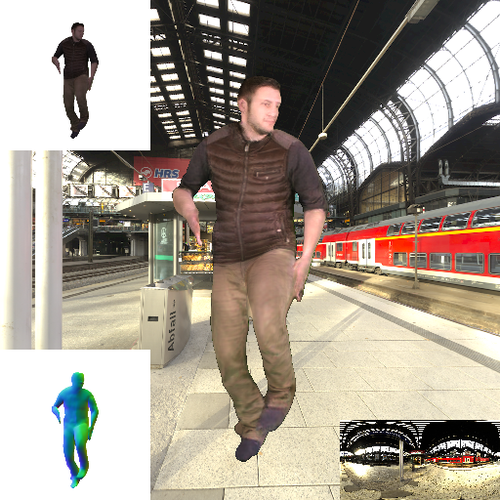} &
    \includegraphics[width=0.2\textwidth]{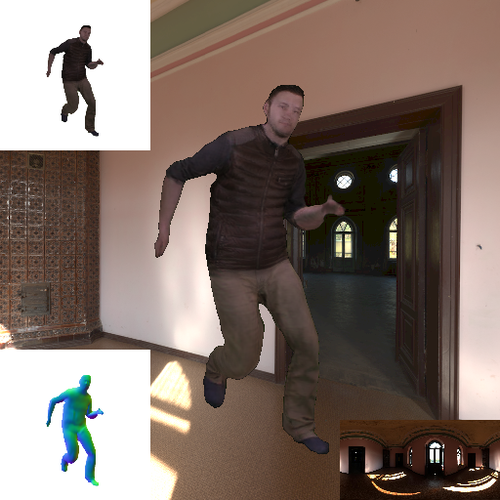}
    \\
    \includegraphics[width=0.2\textwidth]{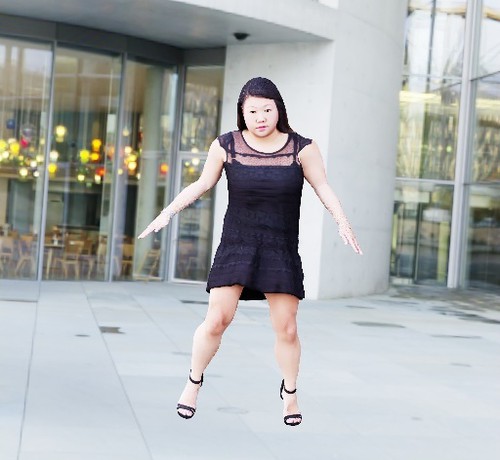} 
    \includegraphics[width=0.2\textwidth]{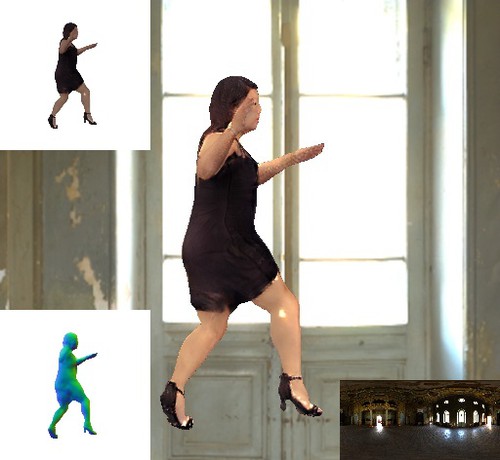} &
    \includegraphics[width=0.2\textwidth]{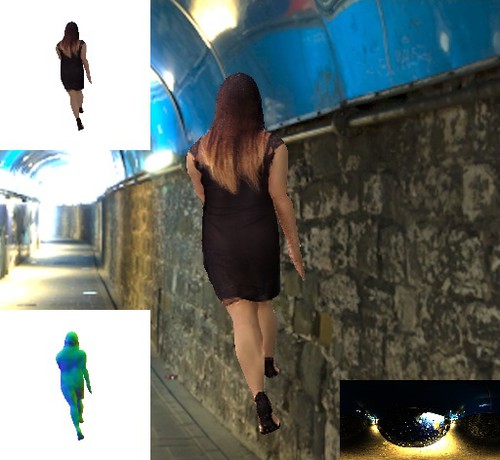} &
    \includegraphics[width=0.2\textwidth]{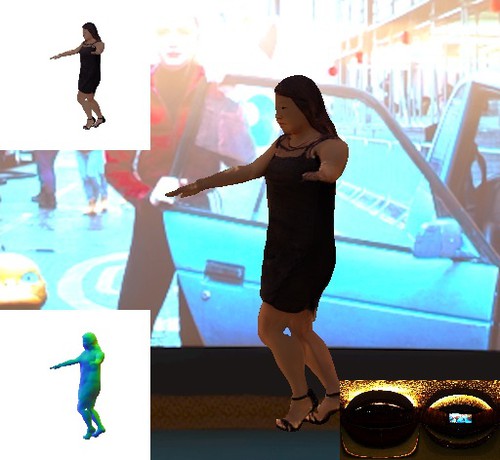} &
    \includegraphics[width=0.2\textwidth]{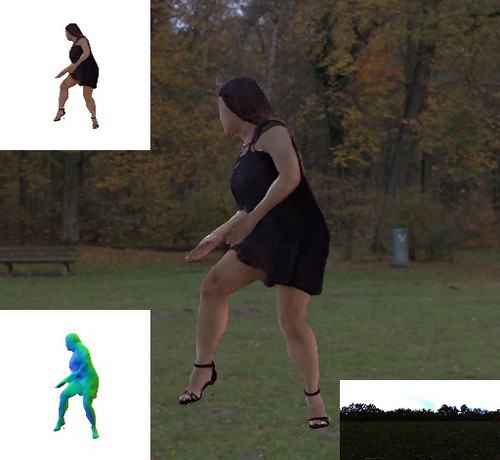}
    \\

  \end{tabular}
}
\vspace{-3mm}
    \captionof{figure}
    {
    We present, RANA, an approach for learning dynamic and relightable full-body avatars from monocular RGB videos. A training frame of the person is shown in the first column. RANA can synthesize images of the person under novel poses, viewpoints, and lighting environments. In the insets, we show the synthesized albedo image, the normal map, and the target HDRI light map. 
    }
    \label{fig:teaser}
\end{center}%
}]
 

%% file: sec/0_abstract.tex
\begin{abstract}
We propose RANA, a relightable and articulated neural avatar for the photorealistic synthesis of humans under arbitrary viewpoints, body poses, and lighting. We only require a short video clip of the person to create the avatar and assume no knowledge about the lighting environment. We present a novel framework to model humans while disentangling their geometry, texture, and also lighting environment from monocular RGB videos. To simplify this  otherwise ill-posed task we first estimate the coarse geometry and texture of the person via SMPL+D model fitting and then learn an articulated neural representation for photorealistic image generation. RANA first generates the normal and albedo maps of the person in any given target body pose and then uses spherical harmonics lighting to generate the shaded image in the target lighting environment. We also propose to pretrain RANA using synthetic images and demonstrate that it leads to better disentanglement between geometry and texture while also improving robustness to novel body poses.  Finally, we also present a new photorealistic synthetic dataset, Relighting Humans, to quantitatively evaluate the performance of the proposed approach.
\end{abstract}

%% file: sec/1_introduction.tex
\section{Introduction}
\label{sec:intro}

Articulated neural avatars of humans have numerous applications across telepresence, animation, and visual content creation. To enable the widespread adoption of these neural avatars, they should be easy to generate and animate under novel poses and viewpoints, able to render in photo-realistic image quality, and easy to relight in novel environments. Existing methods commonly aim to learn such neural avatars using monocular videos~\cite{raj2020anr,peng2021neural,peng2021animatable, su2021anerf,yoon2022motion,wang2022arah}. While this allows photo-realistic image quality and animation, the synthesized images are always limited to the lighting environment of the training video. Other works directly tackle relighting of human avatars but do not provide control over the body pose~\cite{guo2019relightables, meka2022deeprelightable}. Moreover, these approaches often require multiview images recorded in a Light Stage for training, which is limited to controlled settings only. Some recent methods aim to relight RGB videos of a dynamic human but do not provide control over body pose~\cite{chen2022relighting}. 

In this work, we present the Relightable Articulated Neural Avatar (RANA) method, which allows photo-realistic animation of people under any novel body pose, viewpoint, and lighting environment. To create an avatar,
we only require a short monocular video clip of the person in 
 unconstrained environment, clothing, and body pose. 
During inference, we only require the target novel body pose and the target novel lighting information. 

Learning relightable neural avatars of dynamics humans from monocular RGB videos recorded in unknown environments is a challenging problem. First, it requires modeling the complex human body articulations and geometry. Second, in order to allow relighting under novel environments, the texture, geometry, and illumination information have to be disentangled, which is an ill-posed problem to solve from RGB videos~\cite{barron2020shape}. To address these challenges, we first extract canonical, coarse geometry and texture information from the training frames using a statistical human shape model SMPL+D~\cite{loper2015smpl, Alldieck_2018_CVPR, lazova3dv2019}. We then propose a novel convolutional neural network that is trained on synthetic data to remove the shading information from the coarse texture. We augment the coarse geometry and texture with learnable latent features and pass them to our proposed neural avatar framework, which generates refined normal and albedo maps of the person under the target body pose using two separate convolutional networks.  Given the normal map, albedo map, and lighting information, we generate
the final shaded image using spherical harmonics (SH) lighting~\cite{ramamoorthi2001sh}. 
During training, since the environment lighting is unknown, we jointly optimize it together with the person's appearance and propose novel regularization terms to prevent the leaking of lighting into the albedo texture. We also propose to pre-train the avatar using photo-realistic synthetic data with ground-truth albedo and normal maps. During pretraining, we simultaneously train a single avatar model for multiple subjects while having separate neural features for each subject. This not only improves the generalizability of the neural avatar to novel body poses but also learns to decouple the texture and geometry information. For a new subject, we only learn a new set of neural features and fine-tune the avatar model to capture fine-grained person-specific details. In our experiments, the avatar for a novel subject can be learned within 15k training iterations. 

To the best of our knowledge, RANA is the first method to enable relightable and articulated neural avatars. Hence, in order to quantitatively evaluate the performance of our method, we also propose a novel photo-realistic synthetic dataset, Relighting Humans (RH), with ground truth albedo, normals, and lighting information. The Relighting Humans dataset allows for simultaneous evaluation of the performance in terms of novel pose and novel light synthesis. We also perform a qualitative evaluation of RANA on the People Snapshot dataset~\cite{Alldieck_2018_CVPR} to compare with other baselines. 

Our contributions can be summarized as follows:
\begin{itemize}
    \itemsep0em 
    \item We present, RANA, a novel framework for learning relightable articulated neural avatars from short unconstrained monocular videos. The proposed approach is very easy to train and does not require any knowledge about the environment of the training video. 
    \item Our proposed approach can synthesize photorealistic images of humans under any arbitrary body pose, viewpoint, and lighting. It can also be used for relighting videos of dynamic humans. 
    \item We present a new photo-realistic synthetic dataset for quantitative evaluation and to further the research in this direction. 
\end{itemize}

%% file: sec/2_related.tex
\section{Related work}

\noindent\textbf{Mesh Based Human Avatars.} These methods represent human avatars using a rigged mesh and an associated texture map. Earlier methods captured human avatars using multi-view cameras~\cite{carranza2003free,starck2007surface} or with the help of depth sensors~\cite{bogo2015detailed,zhi2020texmesh}. However, their adoption remained limited due to the constrained hardware requirements. The recent works, therefore, focus on creating the avatars from monocular videos~\cite{Alldieck_2018_CVPR, alldieck2018detailed} or images~\cite{lazova3dv2019, yang2021s3, huang2020arch, He2021ARCHAC, alldieck2022phorhum}. The methods~\cite{Alldieck_2018_CVPR, alldieck2018detailed, lazova3dv2019} use body model fitting to capture the humans, while more recent methods use data-driven implicit functions combined with pixel-aligned features~\cite{saito2020pifuhd} for human reconstruction~\cite{zhi2020texmesh, huang2020arch, yang2021s3, alldieck2022phorhum}. The main limitation of these methods is that the shading information is baked into the texture, therefore, the avatars cannot be rendered with novel lights. PHORUM~\cite{alldieck2022phorhum} is the only exception, however, it creates the avatar from a single image and relies on data-driven priors to hallucinate the occluded regions of the person. Hence, the generated images may not be the true representation of the person. In contrast, our approach uses video data to capture a detailed human representation, while also allowing the rendering of the person in novel lighting. 

\begin{table}[t]
\vspace{-3mm}
\setlength\tabcolsep{8pt} %
\centering
\small
     \begin{tabular}{@{} *{5}{l} }
      \rot{novel view} &
      \rot{novel pose} &
      \rot{generalizable} &
      \rot{relightable} &
     Method \\
     \midrule 
     \OK &        &       &     &   NeuralBody\cite{peng2021neural}, HumanNeRF~\cite{weng2022humannerf} \\
     \OK &  \OK   &      &  &  AnimatableNeRf \cite{peng2021animatable}, NeuMan \cite{jiang2022neuman}  \\
     \OK &  \OK   & \OK  &   & ANR \cite{raj2020anr}, TNA \cite{shysheya2019textured}, StylePeople \cite{grigorev2021stylepeople} \\
     \OK &  &  & \OK     &  Relighting4D \cite{chen2022relighting} \\
     \OK & \OK & \OK & \OK  &  RANA (Ours) \\
     \bottomrule
     \end{tabular}
     \vspace{-3mm}
     \caption{Comparison of some of the representative methods for neural human avatar creation. Ours (RANA) is the only method that allows novel view, pose and light synthesis. \vspace{-5mm} 
     }
\label{tab:relwork}
\end{table}

\vspace{2mm}
\noindent\textbf{Neural Human Avatars.} 
More recent methods learn a neural representation of the person and use neural renderers~\cite{tewari2022advances} to directly generate photorealistic images in the target body pose and viewpoints~\cite{balakrishnan2018synthesizing,esser2018variational,chan2019everybody,wang2021dance,huang2021few,yang2021towards}. These methods are generally classified into 2D or 3D neural rendering based methods~\cite{tewari2022advances}. The 3D neural rendering  methods represent the person using neural radiance fields \cite{mildenhall2020nerf} and render the target images using volume rendering~\cite{peng2021neural,weng2022humannerf,jiang2022neuman, peng2021animatable, wang2022arah}. The 2D neural rendering methods, on the other hand, use CNNs to render the images~\cite{raj2020anr,yoon2022motion,grigorev2021stylepeople,zhao2022high}. One limitation of the 3D neural rending methods is that the avatar has to be created from scratch for each person. In contrast, the 2D based methods offer some generalizability  by sharing the neural renderer across multiple subjects~\cite{raj2020anr}. Our method falls into the 2D neural rendering paradigm as we use CNNs to generate the albedo and normal images of the person. In particular, we take inspiration from ANR~\cite{raj2020anr} for designing our framework. Tab.~\ref{tab:relwork} compares existing neural avatar creation methods. Ours is the only method that allows synthesis under novel poses, viewpoints, and lighting, while also being generalizable. 

\vspace{2mm}
\noindent\textbf{Human Relighting.} Relighting of human images has been studied extensively in the literature~\cite{debevec2000acquiring, NeuralFace2017, kanamori_relight2018, sun2019single, zhou2019dpr, wang2020single, tajimaPG21, ji2022geometry, li2013capturing,Lagunas2021humanrelighting,zhang2021neural}. However, these methods cannot render relighted images in novel body poses and viewpoints. Some recent methods allow relighting from novel views but provide no control over the body pose~\cite{guo2019relightables, pandey2021total, chen2022relighting,yeh2022learning}. Our approach, in contrast, provides full control over the body pose, viewpoint, and lighting.

\input{figures/fig_overview.tex}

%% file: figures/fig_overview.tex
\begin{figure*}
    \vspace{-3mm}
    \centering
    \includegraphics[trim={0cm 0cm 0cm 0cm},clip]{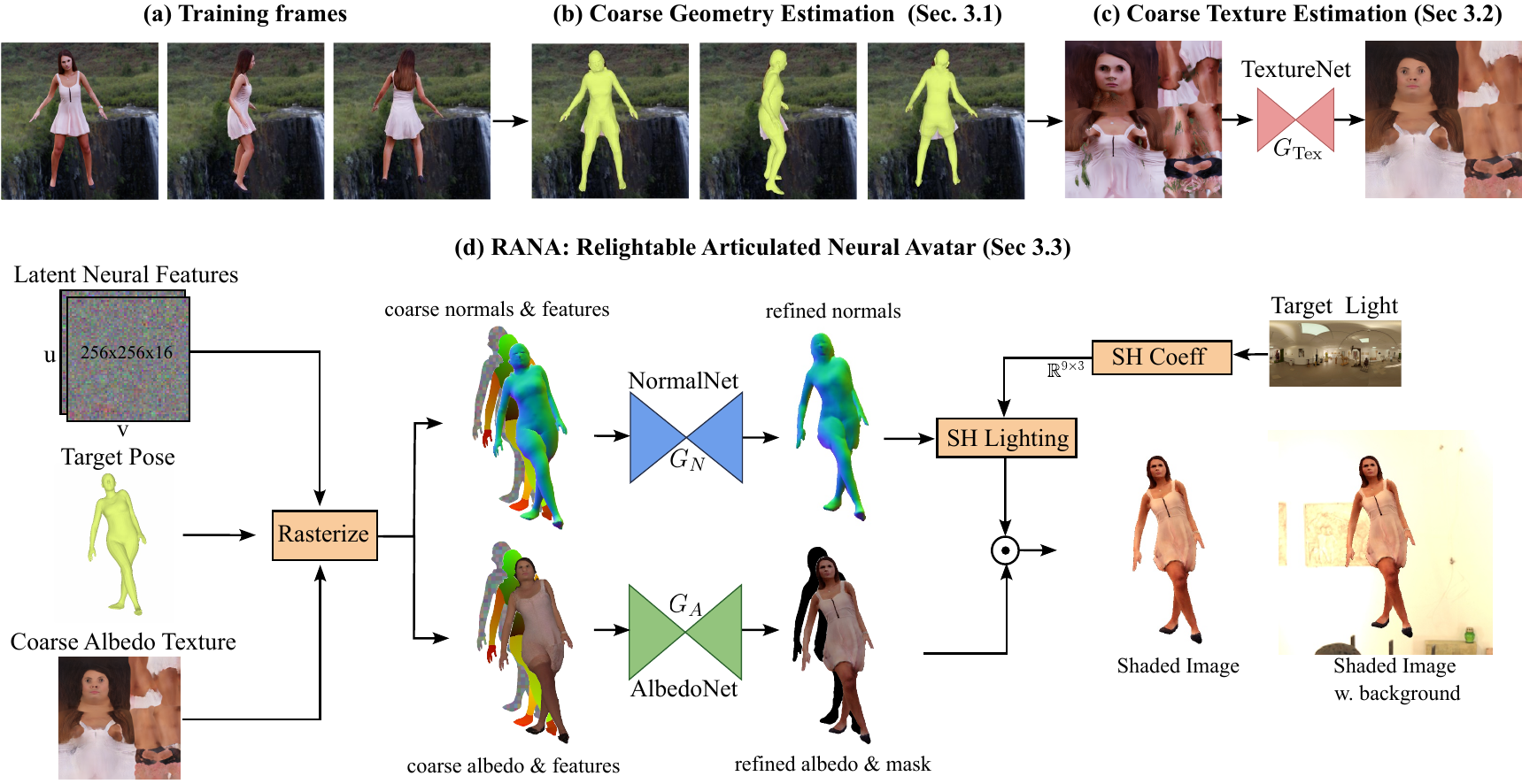}
    \vspace{-7mm}
    \captionof{figure}
    {
    Overview of the proposed approach. (a) shows some training frames. (b) We estimate the coarse geometry of the person using the SMPL+D body model. (c) The SMPL+D fits are used to extract a UV texture map, which we process using TextureNet to obtain a coarse albedo texture map. (d) Given a target body pose, we rasterize person-specific neural features, coarse albedo, and coarse normals from SMPL+D to the target body pose and pass them to NormalNet and AlbedoNet to obtain refined normal and albedo maps, respectively. We then use the normal map and spherical harmonics lighting to obtain the shading image, which is multiplied with refined albedo to produce the shaded image. AlbedoNet also generates a binary mask, which we use to overlay the shaded image onto the background.
    }
    \vspace{-5mm}
    \label{fig:overview}
\end{figure*}%

%% file: sec/3_method.tex
\section{Method}
Our goal is to learn a relightable articulated neural avatar, RANA, that can synthesize photorealistic images of the person under any target body pose, viewpoint, and lighting. To create the avatar,
we use a small video sequence $\mathbf{I}{=}\{I_\mathrm{f}\}_{f=1}^F$ with $F$ video frames and assume no knowledge about the lighting environment and body poses present in the video.  We parameterize RANA using the SMPL~\cite{loper2015smpl} body model to control the animation and use Spherical Harmonics (SH)~\cite{ramamoorthi2001sh}  lighting to model the lighting. During inference, we only need the target body pose and target lighting information in the form of SH coefficients and do not require any exemplar images for groundtruth. Learning RANA from a monocular video requires capturing the geometry and appearance of the dynamic human while also disentangling the shading information. In order to tackle this ill-posed problem, we first capture the coarse geometry using the SMPL+D fits (Sec.~\ref{sec:coarse_geometry}). We use the coarse geometry to extract a coarse texture map from the training images which is converted to an albedo texture map using a convolutional network (Sec.~\ref{sec:coarse_albedo}). We then propose RANA that generates the refined albedo and normal maps. The refined normal maps are used to obtain the shading map using SH lighting which is combined with the refined albedo map to obtain the final image in the target body pose and light (Sec.~\ref{sec:rana}). An overview of our method can be seen in Fig.~\ref{fig:overview}. In the following, we describe each of these modules in greater detail. 

\subsection{Coarse Geometry Estimation}
\label{sec:coarse_geometry}
Given the training frames, we first estimate the coarse geometry of the person including the clothing and hair details. For this, we employ the SMPL+D~\cite{alldieck2018detailed,lazova3dv2019} variant of the SMPL body model~\cite{loper2015smpl}. SMPL is a linear function  $M(\theta, \beta)$ that takes the body pose $\theta \in \mathbb{R}^{72}$ and shape parameters $\beta \in \mathbb{R}^{10}$ as input and produces a triangulated mesh $\mathbf{M} \in \mathbb{R}^{V\times3}$ with $V{=}6890$ vertices. SMPL only captures the undressed shape of the body and ignores the clothing and hair details. For this, SMPL+D adds a set of 3D offsets $\mathbf{D} \in \mathbf{R}^{V{\times}3}$ to SMPL to capture the additional geometric details, \ie, $M(\theta, \beta, \mathbf{D}) \in \mathbb{R}^{V{\times}3}$  can also model clothed humans. We refer the readers to~\cite{lazova3dv2019, alldieck2018detailed} for a detailed description of SMPL+D. 

For fitting SMPL+D to training images, we first estimate the parameters of SMPL using an off-the-shelf method SMPLify3D~\cite{iqbal2021kama}. Since the person in the video remains the same, we optimize a single $\beta$ for the entire video. We then fix the pose $\{\theta\}_{f{=}1}^{F}$  
and shape $\beta$ parameters and optimize for the offsets $\mathbf{D}$ using the following objective:
\begin{equation}
\mathbf{D} = \argmin_{\mathbf{D}'} \sum_{f=1}^{F} \mathcal{L}(M(\theta_f, \beta, \mathbf{D}')) = \mathcal{L}_{\mathrm{Sil}} + \mathcal{L}_{\mathrm{smooth}}.  
\end{equation}
Here the term $\mathcal{L}_{\mathrm{Sil}}$ is the silhouette loss.  We obtain the silhouette of SMPL+D vertices $S_f$ for frame $f$ using a differentiable renderer~\cite{liu2019softras} while the target silhouette $\hat{S}_t$ is obtained using a person segmentation model~\cite{check2017deeplab}. We define the $\mathcal{L}_{\mathrm{Sil}}$ loss as 
\begin{align}
     \mathcal{L}_{\mathrm{Sil}} = \frac{1}{F}\sum_{f=1}^{F} |S_f - \hat{S}_f|. 
\end{align} 
The term $\mathcal{L}_{\mathrm{smooth}}$ is a laplacian smoothing term to encourage the smooth surface of the mesh:
\begin{align}
    \mathcal{L}_{\mathrm{smooth}} = \frac{1}{F}\sum_{f=1}^{F} ||L\mathbf{M}_f||,
\end{align}
where $L$ is the mesh Laplacian operator  
Note that we optimize a single set of $\mathbf{D}$ for the entire video, hence it does not model any pose-dependent geometric deformations. Some examples of SMPL+D can be seen in Fig.~\ref{fig:overview}\textcolor{red}{b}.

\subsection{Coarse Albedo Estimation}
\label{sec:coarse_albedo}
Given the SMPL+D fits for the training frames, we estimate an albedo texture map $T_A$ of the person in the UV space of SMPL. We follow~\cite{Alldieck_2018_CVPR} and first extract a partial texture map for each frame by back-projecting the image colors of all visible vertices to the UV space. The final texture map $T_I$ is then generated by calculating the median color value of most orthogonal texels from all frames. Depending on the available body poses in the training video, the obtained texture map can be noisy, and still have missing regions, \eg, hand regions are often very noisy as no hand tracking is performed during SMPL fitting. Also, to ensure plausible relighting, the unknown shading from the texture map has to be removed, which is a challenging problem since decomposing shading and albedo is an ill-posed problem. 

To address these problems, we propose TextureNet, $G_\mathrm{Tex}$ (Fig.~\ref{fig:overview}\textcolor{red}{c}), which takes a noisy texture map $T_I$ with unknown lighting as input and produces a clean albedo texture map as output, \ie, $T_A = G_{\mathrm{Tex}}(T_I)$. One main challenge for training such a model is the availability of training pairs of noisy/shaded and albedo texture maps. We generate these pairs using 400 rigged characters from the RenderPeople dataset~\cite{renderpeople}. Since each character in RenderPeople has different UV coordinates, we follow~\cite{lazova3dv2019} and register the characters with SMPL to obtain ground-truth UV maps in consistent SMPL UV coordinates. For noisy pairs, we generate images with random poses and lighting and extract texture maps like any other video mentioned above. We train $G_{\mathrm{Tex}}$ using the following losses:
\begin{align}
\mathcal{L}_\mathrm{Tex} = \mathcal{L}_\mathrm{pixel} + \mathcal{L}_{\mathrm{VGG}} + \mathcal{L}_\mathrm{GAN}.  
\end{align}
Here  $\mathcal{L}_\mathrm{pixel}$ is the $L_1$ loss between the predicted and ground-truth albedo texture maps, $\mathcal{L}_\mathrm{VGG}$ is $L_1$ difference between their VGG features, and $\mathcal{L}_{\mathrm{GAN}}$ is a typical GAN loss~\cite{goodfellow2014gan, park2019semantic}. More details about data generation and training of $G_\mathrm{Tex}$ are provided in Sec.~\ref{sec:imp_texture_net}. Some examples of estimated albedo maps can be seen in Fig.~\ref{fig:albedo_texture_maps}. 

\input{figures/fig_albedo_comparison.tex}

\subsection{Relightable Articulated Neural Avatar}
\label{sec:rana}
The coarse albedo texture and geometry obtained so far lack photo-realism and fine-grained details of the person. First, the topology of SMPL+D is fixed and cannot fully capture the fine geometric details, for example, loose clothing or long hairs. Second, the TextureNet can confuse the texture of the person with shading and may remove some  texture details while estimating the albedo texture map. In this section, we present RANA which utilizes the coarse geometry and albedo map and generates photo-realistic images of the person. We parametrize RANA using the SMPL body model and SH lighting~\cite{Alldieck_2018_CVPR}. Specifically, RANA takes the target body pose $\theta$ and the target lighting in the form of second-order spherical harmonics coefficients $E \in \mathbb{R}^{9\times3}$ as input and synthesizes the target image $I^{\theta, E}$ as:
\begin{align}
I^{\theta, E} = \texttt{RANA}(\theta, E, K),
\end{align}
where $K$ corresponds to the intrinsic parameters of the target camera viewpoint. One main challenge in learning such a neural avatar from a short RGB video is to maintain the disentanglement of geometry, albedo, and lighting, as any learnable parameters can overfit the training frames disregarding plausible disentanglement. 
Hence, we design RANA such that a plausible disentanglement is encouraged during training. Specifically, RANA consists of two convolutional neural networks NormalNet, $G_\mathrm{N}$, and AlbedoNet, $G_\mathrm{A}$, each responsible for generating the normal map $I^\theta_N \in \mathbb{R}^{h \times w \times 3}$, and albedo map, $I^\theta_A \in \mathbb{R}^{h \times w \times 3}$, of the person in the body pose $\theta$, respectively. It also consists of a set of subject-specific latent neural features $Z \in \mathbb{R}^{256 \times 256 \times 16}$ in UV coordinates to augment the details available in the coarse albedo map and geometry. 

More specifically, given the target body pose $\theta$, we first generate the SMPL+D mesh $\mathbf{M}^\theta = M(\theta, \beta, \mathbf{D})$, where the shape parameters $\beta$ and clothing offsets $\mathbf{D}$ are the ones obtained in Sec.~\ref{sec:coarse_geometry}. We then use $\mathbf{M}^\theta$ to differentiably rasterize~\cite{liu2019softras} the latent features $Z$ and coarse albedo texture $T_A$ to obtain a features image $I^\theta_Z$ and coarse albedo image $\bar{I}^\theta_A$ in the target body pose. We also rasterize a coarse normal image $\bar{I}^\theta_N$ and a UV image $I^\theta_{\mathrm{uv}}$ using the normals and UV coordinates of $M^\theta$, respectively. The refined normal image $I^\theta_N$ and refined albedo image $I^\theta_A$ are then obtained as 
\begin{align}
    I^\theta_N =& ~G_N(I^\theta_Z, \bar{I}^\theta_N, \gamma(I^\theta_{\mathrm{uv}})),  \\
    I^\theta_A, S^\theta =& ~G_A(I^\theta_Z, \bar{I}^\theta_A, \gamma(I^\theta_{\mathrm{uv}})), 
\end{align}
where $S^\theta$ is the person mask in the target pose and $\gamma$ corresponds to the positional encoding of the UV coordinates~\cite{mildenhall2020nerf}. Given the lighting $E$, we obtain the shading image $I^{\theta,E}_S$ using the normal map $I^\theta_N$ and SH lighting~\cite{ramamoorthi2001sh}. Under the usual assumptions of Lambertian material, distant lighting, and no cast shadows, the final shaded image $I^{\theta,E}$ is then obtained as
\begin{align}
I^{\theta,E} = I_A^\theta \cdot I_S^{\theta,E}. 
\end{align}
An overview of RANA can be seen in Fig.~\ref{fig:overview}\textcolor{red}{d}. Since the lighting environment of the training video is unknown, we also optimize the second order SH coefficients $E \in \mathbb{R}^{9 \times 3}$~\cite{ramamoorthi2001sh} of the training video during training. Note that none of the learnable parameters in RANA depend on the lighting information. Hence, if the disentanglement of normals, albedo, and lighting during training is correct, we can simply replace $E$ during inference with any other novel lighting environment to obtain relit images. 
We train RANA with the following objective: 
\begin{align}
    \mathcal{L} =  \mathcal{L}_\mathrm{pixel} &+ \mathcal{L}_\mathrm{face} + 
    \mathcal{L}_\mathrm{mask} + \mathcal{L}_\mathrm{VGG} + \mathcal{L}_\mathrm{GAN} \\  
    &+ \mathcal{L}_\mathrm{reg}^{\mathrm{albedo}} + \mathcal{L}_\mathrm{reg}^{\mathrm{normal}}. 
    \label{eq:rana_loss}
\end{align}
Here $\mathcal{L}_\mathrm{pixel}$ is the $L_1$ difference between the generated image $I^{\theta, E}$ and the ground-truth training frame, $\mathcal{L}_{\mathrm{face}}$ is the $L_1$ difference between their face regions to assign a higher weight to face, and $\mathcal{L}_\mathrm{mask}$ is the binary-cross-entropy loss between the estimated mask $S^\theta$ using $G_A$ and the pseudo-ground-truth mask obtained using a person segmentation model~\cite{check2017deeplab}. The term $\mathcal{L}_\mathrm{VGG}$ is the $L_1$ difference between the VGG features of generated and ground-truth images, and $\mathcal{L}_\mathrm{GAN}$ is the commonly used GAN loss~\cite{park2019semantic, goodfellow2014gan}. The term 
 $\mathcal{L}_\mathrm{reg}^\mathrm{albedo}$ is the albedo regularization term  that prevents the light information from leaking into the albedo image:
\begin{align}
    \mathcal{L}^{\mathrm{albedo}}_\mathrm{reg} = ||\sigma(I^\theta_A, k) - \sigma(\bar{I}^\theta_A, k)||^2.
\end{align}
Here $I^\theta_A$ is the albedo image obtained using $G_A$, $\bar{I}^\theta_A$ is the coarse albedo image, and $\sigma$ is the Gaussian smoothing operator with a kernel size $k{=}51$. $\mathcal{L}^\mathrm{albedo}_\mathrm{reg}$ encourages the overall color information in $I^\theta_A$ to be close to $\bar{I}^\theta_A$ while disregarding the texture information. Similarly, $\mathcal{L}^{\mathrm{normal}}_\mathrm{reg}$ is the normal regularization loss which prevents the normal image $I^\theta_N$ to move very far from the coarse normal image $\bar{I}^\theta_N$:
\begin{equation}
    \mathcal{L}_\mathrm{reg}^\mathrm{albedo} = |S^\theta_\textrm{smpl}I^\theta_N - S^\theta_\textrm{smpl}\bar{I}^\theta_N|, 
\end{equation}
where $S^\theta_\textrm{smpl}$ is the rasterized mask of SMPL+D mesh. It ensures that the regularization is applied only on the pixels where SMPL+D normals are valid. Note that no ground-truth supervision is provided to $G_A$ and $G_N$. They are mostly learned via the image reconstruction losses, while the disentanglement of normals and albedo is ensured via the novel design of RANA and regularization losses. 

\subsubsection{Pre-training using synthetic data}
\label{sec:pretraining}
While we design RANA such that it generalizes well to novel body poses, the networks $G_A$ and $G_N$ may still overfit to the body poses available in the training video, in particular, when the coarse geometry and albedo are noisy. A significant advantage of RANA is that it can be trained simultaneously for multiple subjects, \ie, we use different neural features $Z$ for each subject while sharing the networks $G_A$ and $G_N$. This not only allows the model to see diverse body poses during pre-training but also helps in learning to disentangle normals and albedo. Hence, we propose to pretrain $G_A$ and $G_N$ on synthetic data. For this, we use 400 rigged characters from the RenderPeople dataset. We generated $150$ albedo and normal images for each subject under random body poses and pretrain both networks using ground-truth albedo and normal images. We use the $L_1$ loss for both terms. For a new subject, we learn the neural features $Z$ from scratch and only fine-tune $G_A$. During our experiments, we found that fine-tuning $G_N$ is not required if the model is pretrained (see Sec.~\ref{sec:ablation}). 

%% file: figures/fig_albedo_comparison.tex
\begin{figure*}[!]
  \centering
  \small
    \setlength{\tabcolsep}{1pt}
    \renewcommand{\arraystretch}{0.4}
\scalebox{0.96}{
  \begin{tabular}{cccccccc}
    \includegraphics[trim={0cm 0cm 9cm 0cm},clip, width=0.155\linewidth]{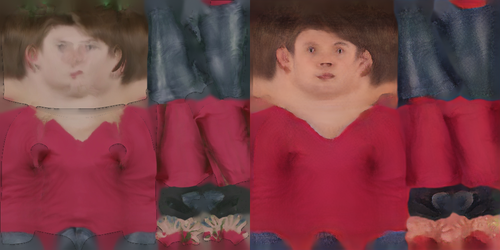} &
    \includegraphics[trim={9cm 0cm 0cm 0cm},clip, width=0.155\linewidth]{figures/albedo_uv_maps/tex-female-4-casual_comparison.png} &
    \includegraphics[trim={0cm 0cm 9cm 0cm},clip, width=0.155\linewidth]{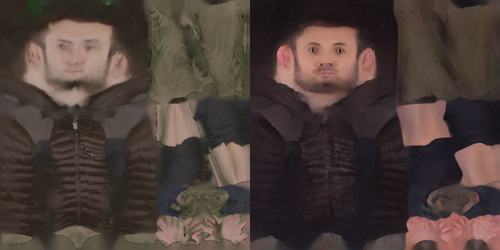} &
    \includegraphics[trim={9cm 0cm 0cm 0cm},clip, width=0.155\linewidth]{figures/albedo_uv_maps/tex-male-2-casual_comparison.png} &
    \includegraphics[trim={0cm 0cm 9cm 0cm},clip, width=0.155\linewidth]{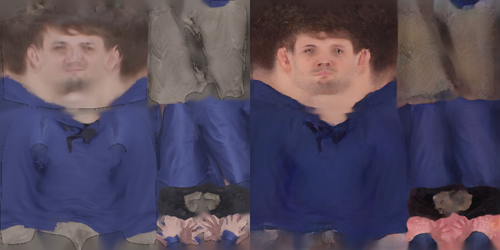} &
    \includegraphics[trim={9cm 0cm 0cm 0cm},clip, width=0.155\linewidth]{figures/albedo_uv_maps/tex-male-9-plaza_comparison.png} \\
    \includegraphics[trim={0cm 0cm 9cm 0cm},clip, width=0.155\linewidth]{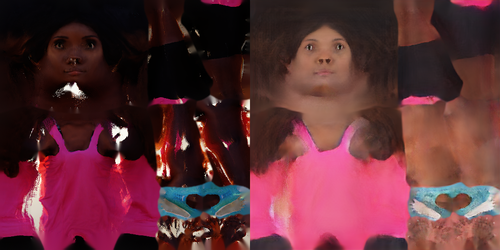} &
    \includegraphics[trim={9cm 0cm 0cm 0cm},clip, width=0.155\linewidth]{figures/albedo_uv_maps/rp_0007_comparison.png} &
    \includegraphics[trim={0cm 0cm 9cm 0cm},clip, width=0.155\linewidth]{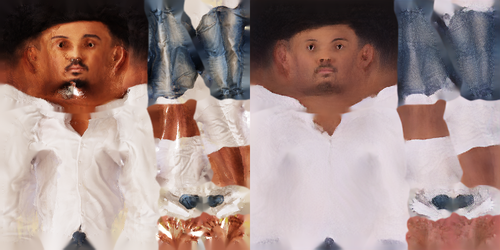} &
    \includegraphics[trim={9cm 0cm 0cm 0cm},clip, width=0.155\linewidth]{figures/albedo_uv_maps/rp_0003_comparison.png} &
    \includegraphics[trim={0cm 0cm 9cm 0cm},clip, width=0.155\linewidth]{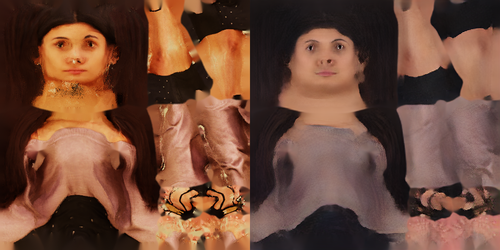} &
    \includegraphics[trim={9cm 0cm 0cm 0cm},clip, width=0.155\linewidth]{figures/albedo_uv_maps/rp_0006_comparison.png} \\
    Noisy/Shaded Map & Albedo Map & Noisy/Shaded Map & Albedo Map & Noisy/Shaded Map & Albedo Map \\
    \vspace{-3mm}
  \end{tabular}
}
\caption{Examples of estimated albedo maps from noisy/shaded maps using our proposed TextureNet (Sec.~\ref{sec:coarse_albedo}). The first row shows some examples from the PeopleSnapshot dataset, while the second row shows examples from our proposed RelightingHuman dataset.  \vspace{-5mm}}
\label{fig:albedo_texture_maps}
\end{figure*}

%% file: sec/4_experiments.tex
\input{figures/fig_ablation_study.tex}

\section{Experiments}
In this section, we evaluate the performance of RANA using two different datasets. We perform an ablation study to validate our design choices and also compare our method with state-of-the-art and other baselines. 

\subsection{Datasets}
\paragraph{Relighting Human Dataset.} We propose a new photorealistic synthetic dataset to quantitatively evaluate the performance of our method. We use $49$ rigged characters from the RenderPeople dataset~\cite{renderpeople} to generate photo-realistic images for each subject. We use HDRI maps from PolyHaven~\cite{polyhaven} to illuminate the characters and use the CMU motion capture dataset~\cite{cmu_mocap} to pose the characters. In contrast to our proposed method that uses image-based lighting, we use full Path Tracing to generate the dataset. Hence, it is the closest setting to an in-the-wild video, and any future work that uses a more sophisticated lighting model can be evaluated on this dataset. For a fair evaluation, we ensure that none of the characters is used during the training in Sec~\ref{sec:coarse_albedo} and Sec~\ref{sec:rana}. All testing images come with a ground-truth albedo map, a normal map, a segmentation mask, and light information. For our experiments, we evaluate on all $49$ characters and learn a separate RANA model for each subject.  
We develop two different protocols for evaluation:\\
\noindent \textbf{a) Novel Pose and Light Synthesis.} This protocol evaluates the quality in terms of novel pose and light synthesis. We generate $100$ training images for each subject rotating $360^\circ$ with A-pose in front of the camera with fixed lighting. For testing, we generate $150$ frames for each subject with random body pose and random light in each frame.   \\ 
\noindent \textbf{b) Novel Light Synthesis.} This protocol evaluates the relighting ability of the methods. We generate $150$ frames for train and test sets. The train set is generated with fixed lighting and random body poses. The body poses in the test set are exactly the same as the train sets, but each frame is generated using a different light source. 

\vspace{2mm}
\noindent\textbf{People Snapshot Dataset. \cite{Alldieck_2018_CVPR}.} This dataset consists of real videos of characters rotating in front of the camera. We use this dataset for qualitative evaluation. 

\subsection{Metrics}
We report several metrics to evaluate the quality of synthesized images as well as the disentanglement of normal and albedo images. For synthesized images and albedo maps, we use Learned Perceptual Patch Similarity (LPIPS $\downarrow$)~\cite{zhang2018perceptual},  Deep Image Structure and Texture Similarity (DISTS~$\downarrow$)~\cite{ding2022dists}, Structural Similarity Index (SSIM~$\uparrow$)~\cite{wang2004ssim} and Peak Signal-to-Noise Ratio (PSNR~$\uparrow$). For normals, we compute the error in degrees ($^{\circ}$).

\subsection{Ablation study}
\label{sec:ablation}
We evaluate different design choices of RANA in Tab.~\ref{tab:ablation_study} and Fig~\ref{fig:ablation}. We use protocol-a of the Relighting Humans dataset for all experiments. We first report the results of the final model which includes all loss terms in \eqref{eq:rana_loss} and pretraining using synthetic data~(Sec.~\ref{sec:pretraining}).  The full model achieves an LPIPS score of $0.217$ for image synthesis and $0.219$ for albedo map reconstruction. If we remove the loss term $\mathcal{L}^\mathrm{albedo}_\mathrm{reg}$ from \eqref{eq:rana_loss}, the LPIPS scores for image and albedo map reconstruction increase to $0.249$ and $0.264$, respectively. Note that the error for albedo maps increases significantly while the error for normal maps remains roughly the same. This indicates that without  $\mathcal{L}^\mathrm{albedo}_\mathrm{reg}$ the light information leaks into the albedo image. An example of this behavior can also be seen in Fig~\ref{fig:ablation}\textcolor{red}{c} (w/o $\mathcal{L}^\mathrm{albedo}_\mathrm{reg}$). 

Next, we evaluate the impact of coarse geometry and albedo texture on RANA. If we do not use coarse geometry and albedo, LPIPS score increases to $0.242$ as compared to $0.217$ for the full model. The normal error also increases to $65.9^\circ$ from $64.3^\circ$. This is also evident from the qualitative results shown in Fig~\ref{fig:ablation}\textcolor{red}{d}, indicating that coarse geometry and albedo help in improved image synthesis quality, in particular when the target body pose is far from the training poses. Next, we evaluate the impact of pretraining on synthetic data. Without the pretraining, all error metrics increase significantly. Specifically, the LPIPS score for image reconstruction increases from $0.217$ to $0.301$, while the normal error increases from $64.3^\circ$ to $74.2^\circ$. If we look at Fig~\ref{fig:ablation}\textcolor{red}{e}, we can see that shading information leaks into both the normals and albedo maps. Hence, pretraining the networks also help with the plausible disentanglement of geometry, texture, and light. Thanks to the design of RANA, we can pretrain on as many subjects as available, which is not possible with most of the state-of-the-art methods for human synthesis~\cite{peng2021animatable, peng2021neural, wang2022arah}. Finally, As discussed in Sec~\ref{sec:pretraining}, we keep the network $G_N$ fixed during finetuning if RANA is pretrained on synthetic data. In the last row of Tab.~\ref{tab:ablation_study}, we evaluate the case when $G_N$ is also fine-tuned. We can see that it has a negligible impact on the results.

\input{figures/fig_exp_NP}
\subsection{Comparison with other methods}
Since RANA is the first neural avatar method that allows novel light and pose synthesis, we ourselves build some baselines as follows:\\
\noindent\textbf{SMPL+D:} We rasterize the SMPL+D mesh normals and albedo texture ( Sec.~\ref{sec:coarse_geometry} \& Sec.~\ref{sec:coarse_albedo}) in the target body pose and use SH lighting to generate the shaded images. \\
\noindent\textbf{ANR~\cite{raj2020anr}+RH{\cite{kanamori_relight2018}}:} We train an ANR~\cite{raj2020anr} model which synthesizes images in the lighting of the training video. We then pass the generated images to the single-image human relighting method~\cite{kanamori_relight2018} to obtain the relighted images for the target light. We use the publicly available source code and models of~\cite{kanamori_relight2018}. \\
\noindent\textbf{Relighting4D~\cite{chen2022relighting}} is a state-of-the-art human video relighting method. We use the publicly available source code and train it on our dataset. 

The results are summarized in Tab.~\ref{tab:baselines_and_sota} and  Fig~\ref{fig:qualitatives}. We do not report results of Relighting4D~\cite{chen2022relighting} for protocol-a since it cannot handle novel body poses as can be seen in Fig~\ref{fig:qualitatives} (column-5). For protocol-a, our method clearly outperforms other baselines for final image synthesis results. Surprisingly, the SMPL+D baseline yields better numbers for albedo reconstruction, even though it provides overly smooth albedo textures. Our qualitative investigation (see Sec.~\ref{sec:rana_vs_smpld}) suggests that the used image quality assessment metrics penalize color differences more than missing texture details. For very bright scenes, RANA can still leak some lighting information to the albedo texture resulting in higher errors for albedo maps even though it provides significantly better texture details than SMPL+D. This is evident from Fig.~\ref{fig:qualitatives} and the final image synthesis results where RANA significantly outperforms SMPL+D baseline. 

For protocol-b, RANA outperforms Relighting4D~\cite{chen2022relighting} across all metrics. Some qualitative comparisons can be seen in Fig.~\ref{fig:qualitatives} (columns 6-7), where RANA clearly yields better image relighting results. Note that each model for  Religthing4D~\cite{chen2022relighting} requires 260k iterations for training whereas RANA models are trained only for 15k iterations, thanks to our novel design that allows pre-training on synthetic data, allowing quick fine-tuning for new subjects. In contrast, Relighting4D~\cite{chen2022relighting} by design cannot be pretrained easily on multiple subjects.  Finally, we provide additional qualitative results in the \href{https://youtu.be/s-hIhIMjPqQ}{supplementary video}.

%% file: figures/fig_ablation_study.tex
\begin{figure*}  
\centering
\small
    \begin{tabular}{ccccc}
        \multicolumn{5}{l}{\includegraphics[width=.9\textwidth]{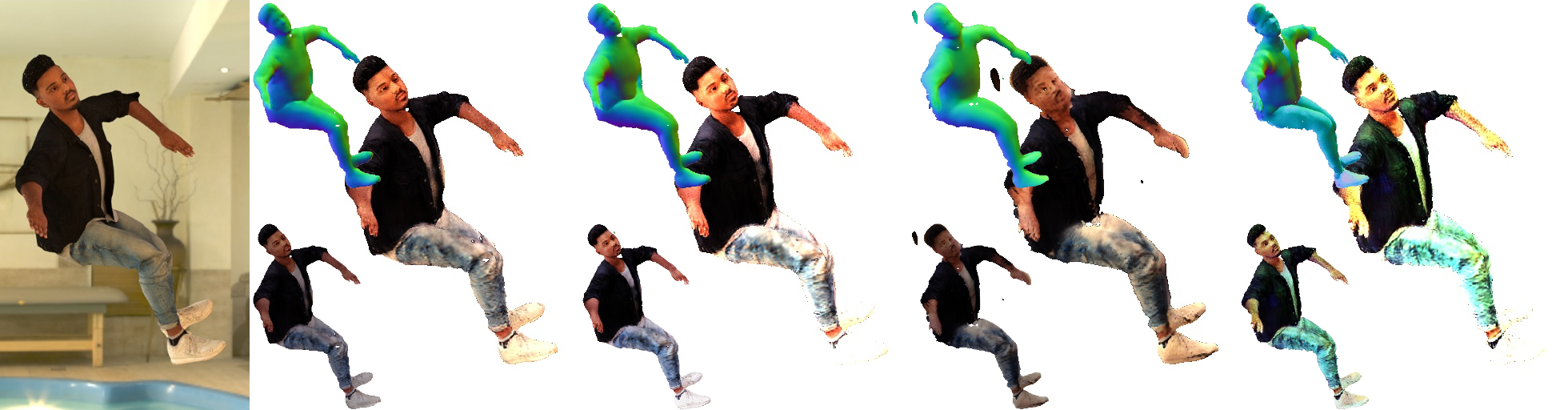}} \\
        ~~(a) Ground Truth &
        ~~~~~(b) Full Model &
        ~~~~~~~~~~~~(c) w/o $\mathcal{L}^\mathrm{albedo}_\mathrm{reg}$ & 
        ~~~~~~~(d) w/o coarse geo. \& tex. & 
        (e) w/o pre-training~~ \\
    \end{tabular}
    \vspace{-3mm}
\caption{\textbf{Ablation Study}. Impact of the different components of the proposed approach. Our full model yields the best results. Without the  $\mathcal{L}^\mathrm{albedo}_\mathrm{reg}$ loss, the light information leaks into the albedo texture resulting in incorrect illumination. If we do not use coarse geometry and albedo texture, the resulting model does not generalize well to novel body poses. Similarly, training the model from scratch, without any pretraining on synthetic data, can result in an incorrect disentanglement of texture and geometry. }
\label{fig:ablation}
\end{figure*}

\begin{table*}[!t]
    \vspace{-3mm}
    \centering
    \small
    \setlength{\tabcolsep}{2mm}{
    \vspace{5pt}
    \renewcommand\arraystretch{0.975}
    \resizebox{0.9\linewidth}{!}{
    \begin{tabular}{lcccc|c|cccc} 
    \toprule
    \multirow{2}{*}{Method}  & 
    \multicolumn{4}{c}{\textbf{Image}} & \multicolumn{1}{c}{\textbf{Normal Map}}  & \multicolumn{4}{c}{\textbf{Albedo Map}}\\ 
    & {LPIPS\;$\downarrow$} &  {FLIP\;$\downarrow$} & {SSIM\;$\uparrow$} & {PSNR\;$\uparrow$} &  {Degree$^\circ\downarrow$} & {LPIPS\;$\downarrow$} &  {FLIP\;$\downarrow$} & {SSIM\;$\uparrow$} & {PSNR\;$\uparrow$} \\  
    \midrule
    Full model &  \bf 0.217  & \bf 0.204  &  \bf 0.751  &  \bf 19.498  &  64.350  &  \bf 0.219  &  \bf 0.207  &  \bf 0.779  &  \bf 21.832 \\
    ~~ w/o $\mathcal{L}_\mathrm{albedo}$ & 0.249  &  0.241  &  0.697  &  15.199  &  64.064  &  0.264  &  0.257  &  0.713  &  15.688 \\
    ~~ w/o coarse geo. \& tex. & 0.242  &  0.222  &  0.730  &  18.580  &  65.887  &  0.260  &  0.232  &  0.751  &  20.778 \\
    ~~ w/o pre-training & 0.301  &  0.293  &  0.669  &  13.798  &  74.215  &  0.327  &  0.307  &  0.696  &  15.632 \\ \midrule 
    ~~ fine-tune $G_N$ & 0.219  &  0.205  &  0.746  &  19.198  & \bf 64.226  &  0.221  &  0.210  &  0.778  &  21.577 \\ 
    \bottomrule
    \end{tabular}}}
    \vspace{-3mm}
    \caption{Ablation study. We evaluate the impact of different components of the proposed method. See Fig.~\ref{fig:ablation} for a qualitative comparison. \vspace{-5mm} } 
    \label{tab:ablation_study}
\end{table*}

%% file: figures/fig_exp_NP.tex
 \begin{figure*}
 \vspace{-3mm}
    \centering
    \small
\scalebox{0.98}{
    \setlength{\tabcolsep}{1pt}
    \renewcommand{\arraystretch}{0.4}
  \begin{tabular}{c|cccc|cc}
      & \multicolumn{4}{
c|}{\bf Protocol-A} &  \multicolumn{2}{|c}{\bf Protocol-B}  \\

    \includegraphics[width=0.14\textwidth]{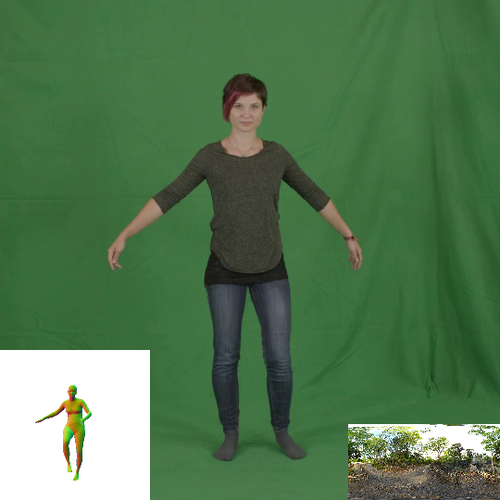} &
    \includegraphics[width=0.14\textwidth]{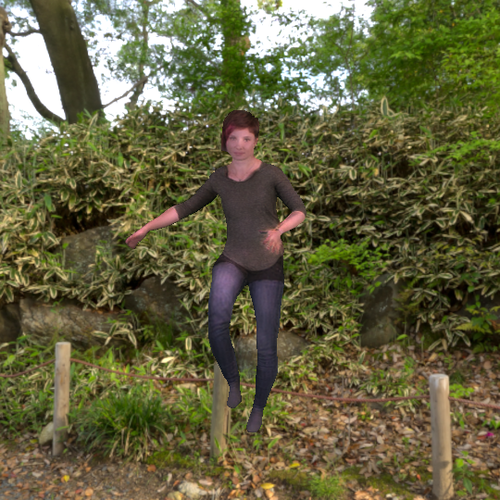} & 
    \includegraphics[width=0.14\textwidth]{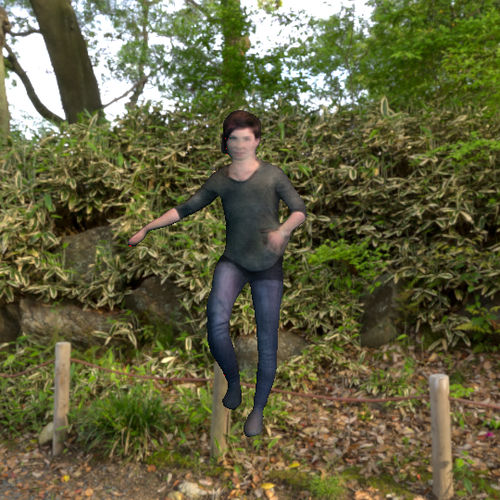} &  
    \includegraphics[width=0.14\textwidth]{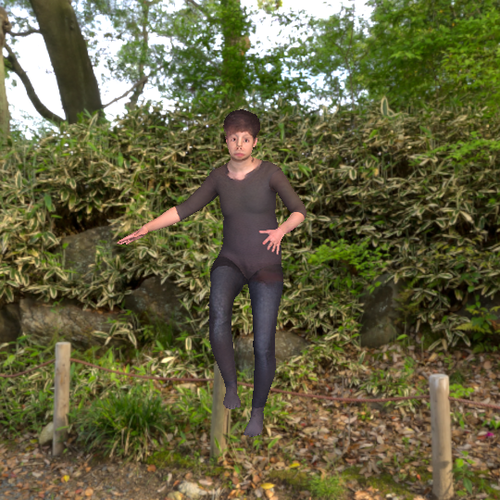} &
    \includegraphics[width=0.14\textwidth]{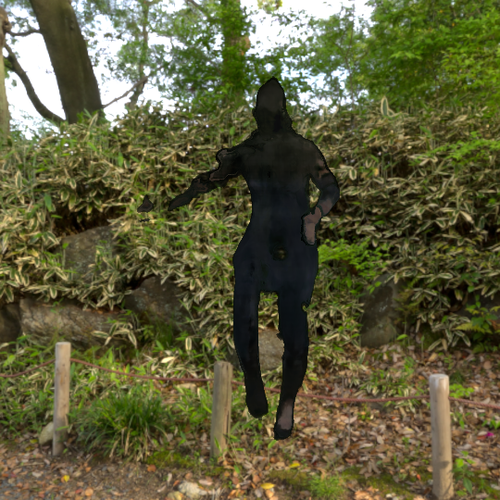} &
    \includegraphics[width=0.14\textwidth]{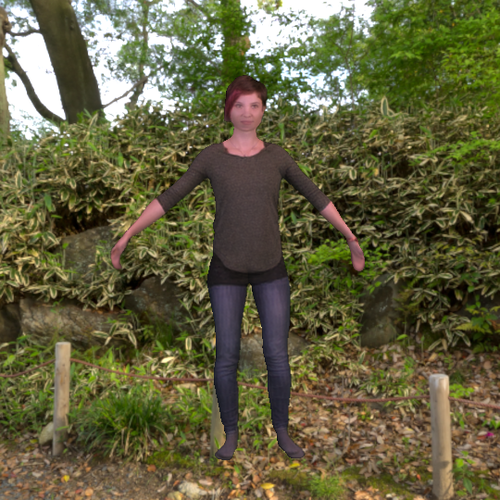} &
    \includegraphics[width=0.14\textwidth]{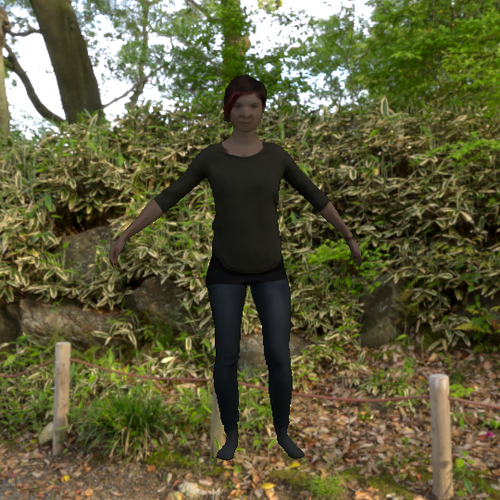}  \\

    \includegraphics[width=0.14\textwidth]{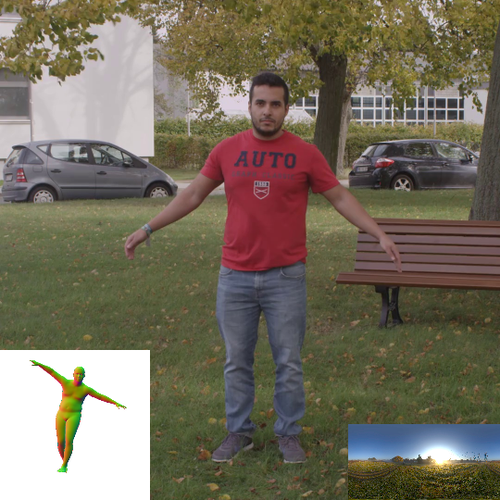} &
    \includegraphics[width=0.14\textwidth]{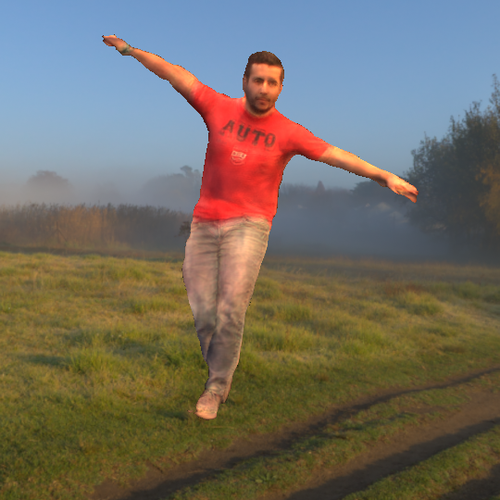} & 
    \includegraphics[width=0.14\textwidth]{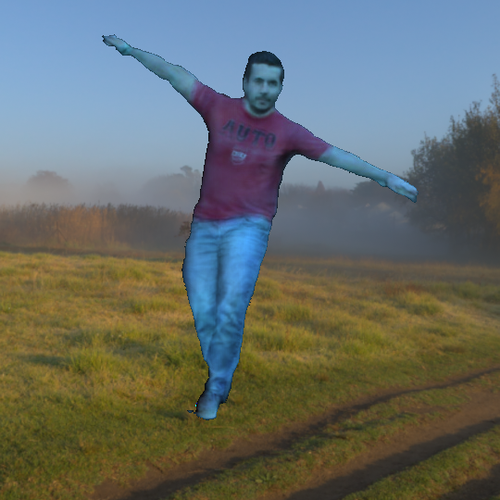} &  
    \includegraphics[width=0.14\textwidth]{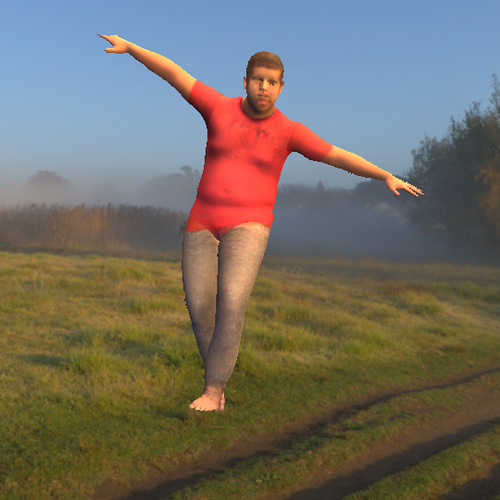} &
    \includegraphics[width=0.14\textwidth]{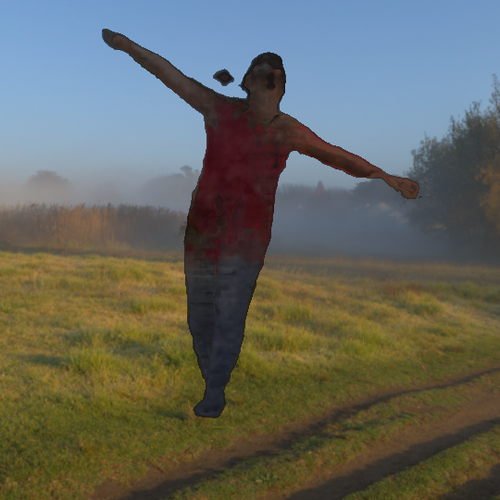} &
    \includegraphics[width=0.14\textwidth]{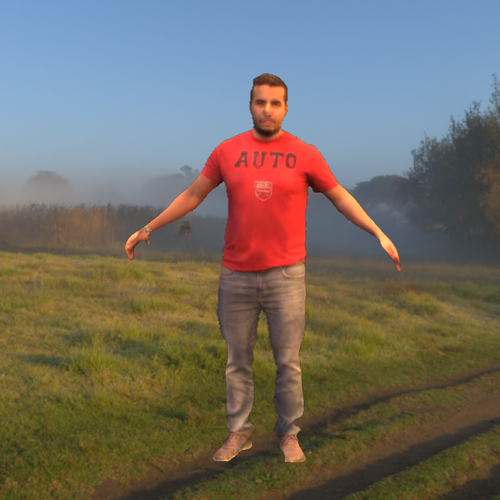} &
    \includegraphics[width=0.14\textwidth]{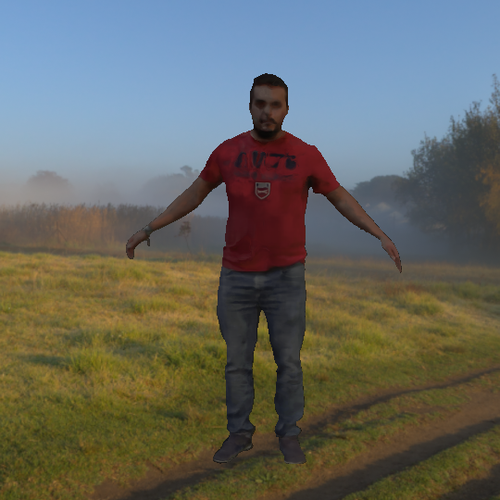}  \\
    
    \includegraphics[width=0.14\textwidth]{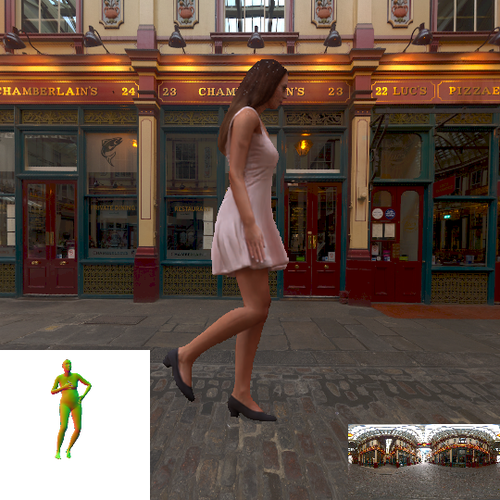} &
    \includegraphics[width=0.14\textwidth]{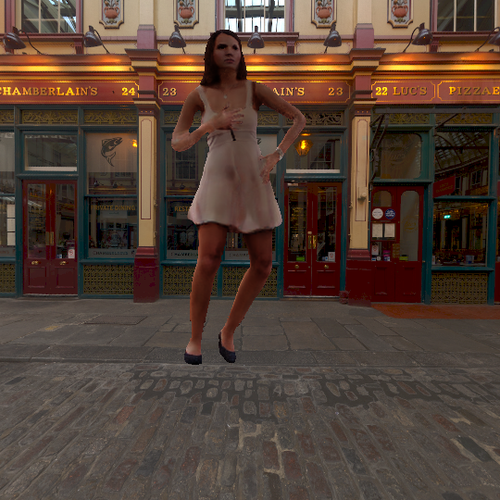} & 
    \includegraphics[width=0.14\textwidth]{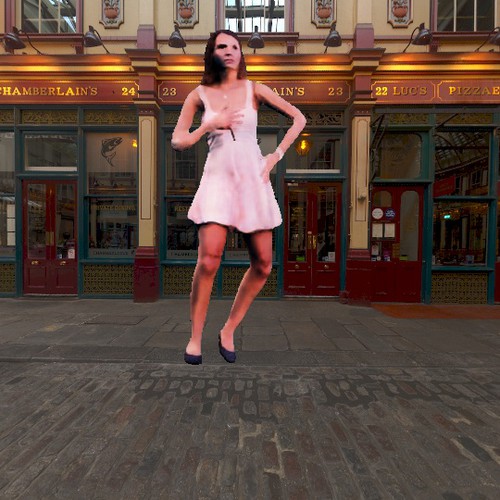} &  
    \includegraphics[width=0.14\textwidth]{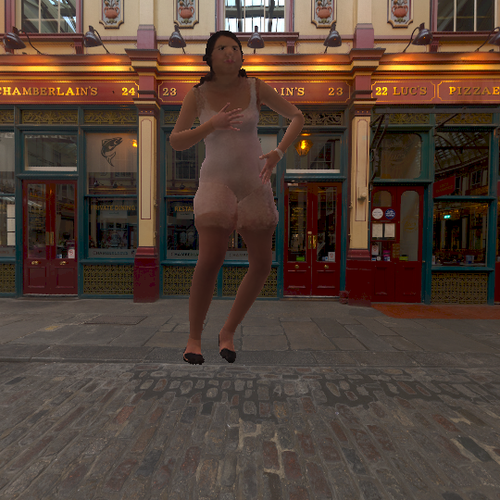} &
    \includegraphics[width=0.14\textwidth]{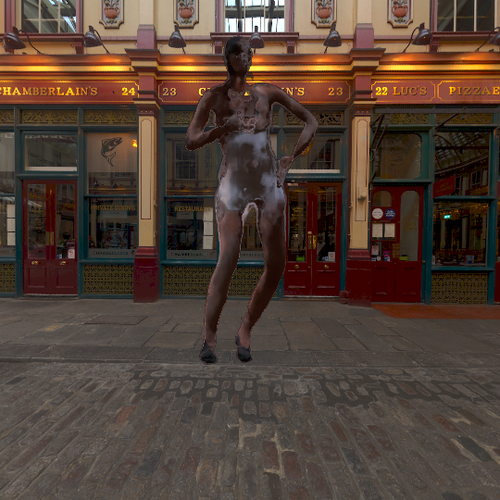} &
    \includegraphics[width=0.14\textwidth]{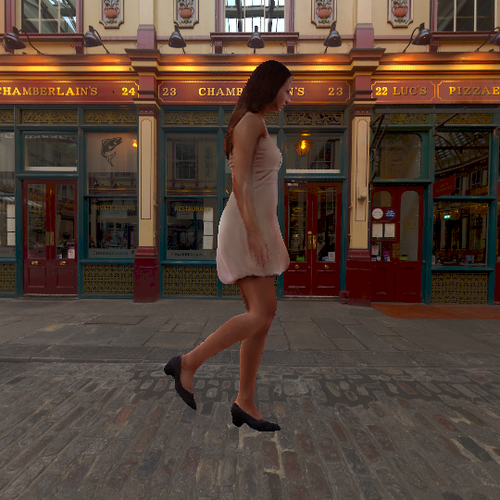} &
    \includegraphics[width=0.14\textwidth]{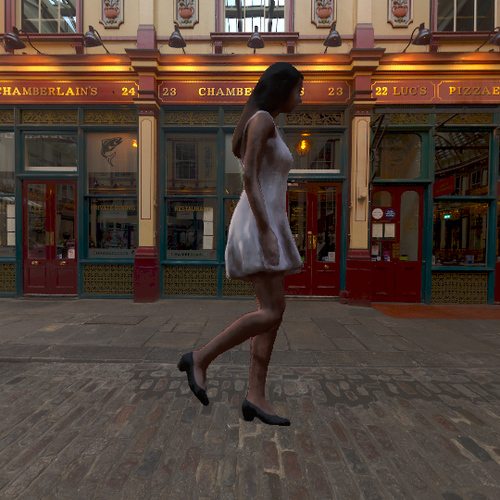}  \\
    
    \includegraphics[width=0.14\textwidth]{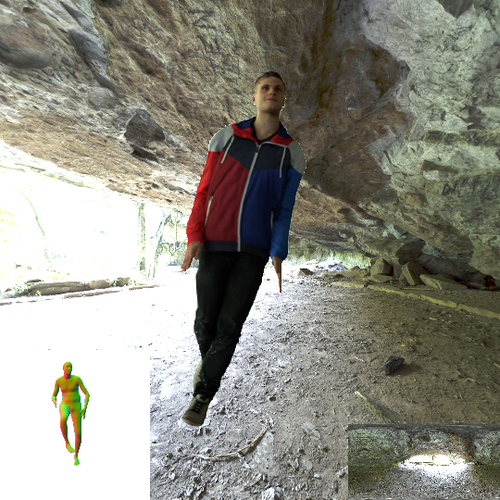} &
    \includegraphics[width=0.14\textwidth]{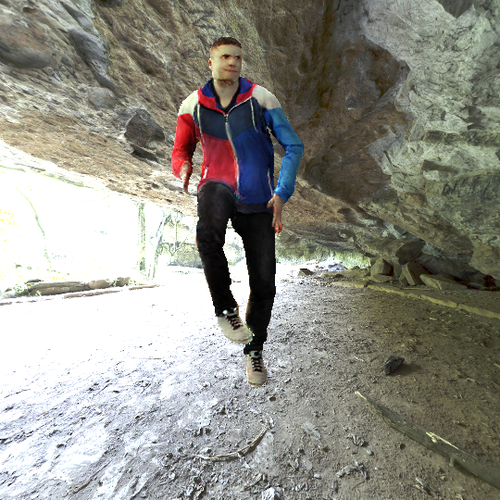} & 
    \includegraphics[width=0.14\textwidth]{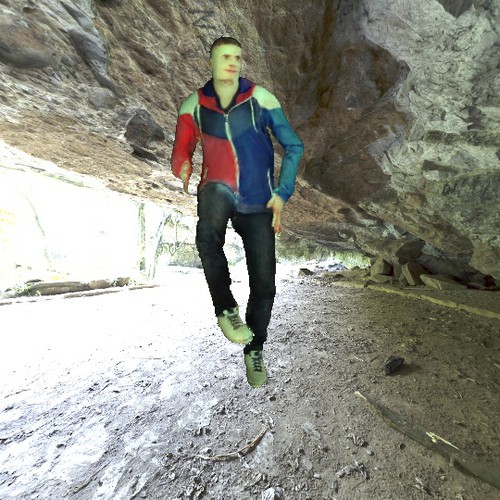} &  
    \includegraphics[width=0.14\textwidth]{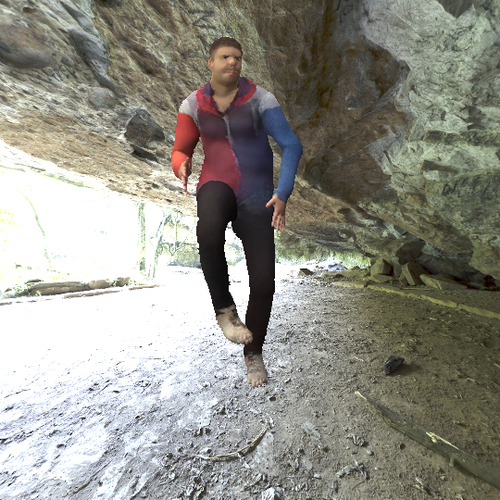} &
    \includegraphics[width=0.14\textwidth]{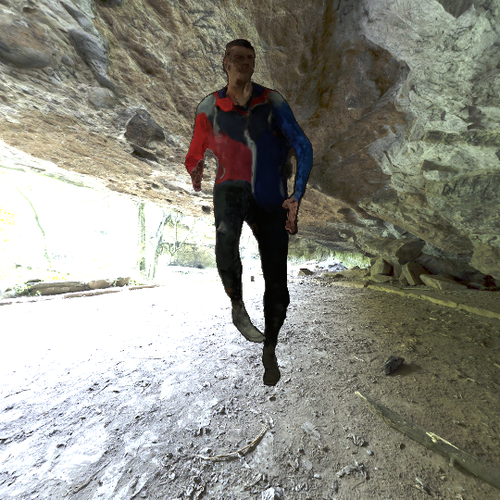} &
    \includegraphics[width=0.14\textwidth]{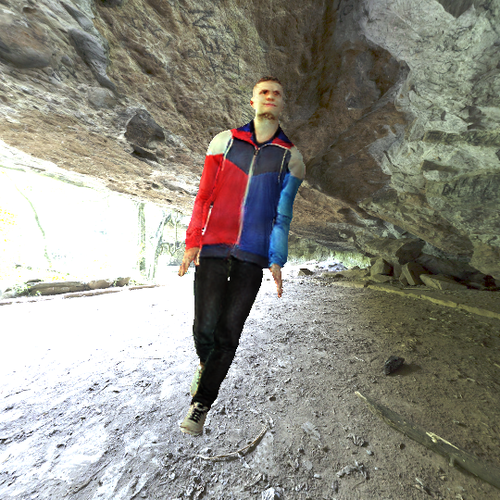} &
    \includegraphics[width=0.14\textwidth]{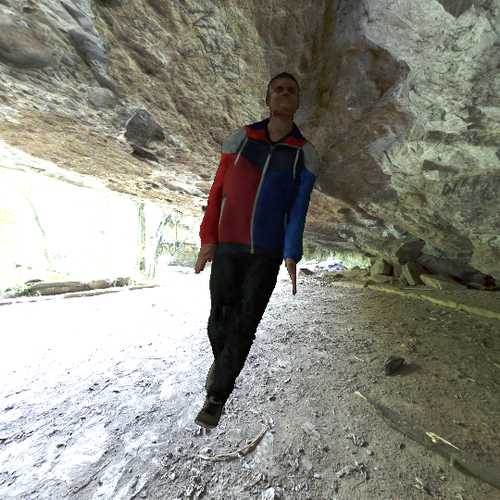}  \\

    Reference & RANA (Ours) & ANR\cite{raj2020anr}+RH\cite{kanamori_relight2018} & SMPL+D & Relighting4D~\cite{chen2022relighting} & RANA (Ours)  & Relighting4D~\cite{chen2022relighting}  \\ 
  \end{tabular}
    }
    \vspace{-3mm}
    \captionof{figure}
    {
    Comparison with the baselines and state-of-the-art methods. Column 1 shows a reference frame with the target body pose and lighting in the insets. In the absence of  true reference images, for the Snapshot dataset (rows 1-2), we show training frames for reference. Columns 2-5 compare different methods for protocol-a, while columns 6-7 provide a comparison for protocol-b. 
    \vspace{-2mm}
    }    
    \label{fig:qualitatives}
\end{figure*}%

 \begin{table*}[t]
 \vspace{-1mm}

    \centering
    \small
    \setlength{\tabcolsep}{2mm}{
    \vspace{5pt}
    \renewcommand\arraystretch{0.975}
    \resizebox{0.9\linewidth}{!}{
    \begin{tabular}{lcccc|c|cccc} 
    \toprule
    \multirow{2}{*}{Method}  & 
    \multicolumn{4}{c}{\textbf{Image}} & \multicolumn{1}{c}{\textbf{Normal Map}}  & \multicolumn{4}{c}{\textbf{Albedo Map}}\\ 
    & {LPIPS\;$\downarrow$} &  {DISTS\;$\downarrow$} & {SSIM\;$\uparrow$} & {PSNR\;$\uparrow$} &  {Degree$^\circ\downarrow$} & {LPIPS\;$\downarrow$} &  {DISTS\;$\downarrow$} & {SSIM\;$\uparrow$} & {PSNR\;$\uparrow$} \\  
    \midrule
    \multicolumn{10}{c}{\bf{Protocol (a):} Novel Pose and Light Synthesis} \\ 
    \midrule
    Ours &  \bf 0.217  &  \bf 0.204  &  \bf 0.751  &  19.498  &  64.350  &  0.219  &  0.207  &  0.779  &  21.832 \\
    SMPL+D & 0.265  &  0.225  &  0.751  &  \bf 19.678  &  \bf 64.121  &  \bf 0.216  & \bf 0.182  &  \bf 0.811  &  \bf 22.623 \\
    ANR~\cite{raj2020anr} + RH~\cite{kanamori2018relighing} & 0.275 & 0.416 & 0.664 & 17.495 & N.A. & 0.266 & 0.429 & 0.656 & 14.804 \\
    \midrule
    \multicolumn{10}{c}{\bf{Protocol (b):} Novel Light Synthesis} \\ 
    \midrule
    Ours &  \textbf{0.173}  &  \textbf{0.171}  &  \textbf{0.842}  &  \textbf{22.338}  &  \textbf{62.823}  &  \textbf{0.200}  &  \textbf{0.179}  &  \textbf{0.865}  &  \textbf{24.721} \\
    Relighting4D~\cite{chen2022relighting} & 0.192 & 0.342 & 0.654 & 21.080 & 65.099 & 0.263 & 0.374 & 0.593 & 20.014  \\
    \bottomrule
    \end{tabular}}}
    \vspace{-2mm}
    \caption{Comparison with the baselines and state-of-the-art methods. See Fig.~\ref{fig:qualitatives} for qualitative comparison. } \vspace{-3mm}
    \label{tab:baselines_and_sota}
\end{table*} 

%% file: sec/5_conclusion.tex
\section{Conclusion and Future Work}
We presented RANA which is a novel framework for learning relightable and articulated neural avatars of humans. We demonstrated that RANA can model humans from unconstrained RGB videos while also disentangling their geometry, albedo texture, and environmental lighting. We showed that it can generate photorealistic images of people under any novel body pose, viewpoints, and  lighting. RANA can be trained simultaneously for multiple people and we showed that pretraining it on multiple (400) synthetic characters significantly improves the image synthesis quality. We also proposed a new photorealistic synthetic dataset to quantitatively evaluate the performance of our proposed method, and believe that it will prove to be very useful to further the research in this direction. 

The most pressing limitation of RANA is the assumption of Lambertian surface, no cast shadows, and image-based lighting.  In the future, we hope to incorporate more sophisticated physically-based rendering in our framework which will hopefully result in better image quality and normal maps with more details. Moreover, RANA does not explicitly model motion-dependent clothing deformations. Modeling clothing deformations from a short video clip would be interesting future work. 

%% file: appendix_content.tex
\section{Appendix}

We provide the implementation details of our proposed approach in Sec.~\ref{sec:imp_texture_net} and Sec.~\ref{sec:impl_rana}. We also provide more details about our proposed Relighting Humans dataset in Sec.~\ref{sec:dataset}. Finally, we provide qualitative results to compare RANA with SMPL+D baseline for albedo texture map estimation in Sec.~\ref{sec:rana_vs_smpld}. The qualitative results augment our comments regarding Tab.~\ref{tab:baselines_and_sota}.

\subsection{Implementation details of TextureNet}
\label{sec:imp_texture_net}

We use a vanilla U-Net architecture for TextureNet. It takes a noisy/shaded UV texture map with a resolution of 512$\times$512 as input and produces the abledo texture with the same resolution as output. We also concatenate a 2D tensor of UV coordinates with the input texture map to provide part-specific information to TextureNet. We train the network using Adam optimizer with a batch size of 8 and a learning rate of $1e^{-4}$ with cosine annealing and a minimum learning rate of $1e^{-5}$. To avoid overfitting during training, we perform random noise augmentation to the input texture maps including coarse dropout, gaussian noise,  random brightness, and MixUp ($\beta=0.4$)~\cite{zhang2018mixup}. As mentioned in the paper we train the model with the following loss function:
\begin{align*}
\mathcal{L}_\mathrm{Tex} = & \mathcal{L}_\mathrm{pixel}(T_A, \hat{T}_A) + \\
                           & \lambda_{\mathrm{VGG}}\mathcal{L}_{\mathrm{VGG}}(T_A, \hat{T}_A) +  \\
                           & \lambda_\mathrm{GAN}\mathcal{L}_\mathrm{GAN}(T_A),
\end{align*}
where $T_A$ and $\hat{T}_A$ are the predicted and ground-truth albedo texture maps, respectively. $\mathcal{L}_\mathrm{pixel}$ and $\mathcal{L}_\mathrm{VGG}$ correspond to the $L_1$ difference between ground-truth and predicted albedo texture maps and their $\mathrm{VGG}$ features, respectively.  We use VGG16 to calculate the VGG features and use the features from \texttt{relu\_1\_2}, $\texttt{relu\_2\_2}$,
$\texttt{relu\_3\_3}$ and $\texttt{relu\_4\_3}$ layers.  For $\mathcal{L}_\mathrm{GAN}$ we use the PatchGAN discriminator~\cite{isola2017image}. We empirically set $\lambda_\mathrm{VGG}{=}1$ and $\lambda_{\mathrm{GAN}}{=}10$.

\subsection{Implementation details of RANA} 
\label{sec:impl_rana}

Similar to TextureNet, we use vanilla U-Net for AlbedoNet and NormalNet. 

\subsubsection{Pretraining.} 
For pretraining RANA, we use 400 characters from RenderPeople and generate 150 samples in the random body poses for each character. Each sample consists of a ground-truth albedo map, a normal map, and the person segmentation mask. We then train AlbedoNet and NormalNet using  Adam optimizer with a batch size of $16$ and learning rate of $1e^{-4}$ with cosine annealing and minimum learning rate of $1e^{-5}$.  We optimize the following objective:
\begin{align*}
\mathcal{L} = \mathcal{L_\mathrm{normal}} + \lambda_\mathrm{a}\mathcal{L_\mathrm{albedo}} + \lambda_m \mathcal{L}_\mathrm{mask}
\end{align*}
where, 
\begin{align*}
\mathcal{L}_\mathrm{normal} = & \mathcal{L}_\mathrm{pixel}(I^\theta_N, \hat{I}^\theta_N) + \\ 
                              & \lambda_{\mathrm{VGG}}\mathcal{L}_{\mathrm{VGG}}(I^\theta_N, \hat{I}^\theta_N) + \\
                              & \lambda_\mathrm{GAN}\mathcal{L}_\mathrm{GAN}(I_N^\theta),  
\end{align*}
\begin{align*}
\mathcal{L}_\mathrm{albedo} = & \mathcal{L}_\mathrm{pixel}(I^\theta_A, \hat{I}^\theta_A) + \\ 
                              & \lambda_{\mathrm{VGG}}\mathcal{L}_{\mathrm{VGG}}(I^\theta_A, \hat{I}^\theta_A) + \\
                              & \lambda_\mathrm{GAN}\mathcal{L}_\mathrm{GAN}(I_A^\theta),  
\end{align*}
and
\begin{align*}
\mathcal{L}_\mathrm{mask} = \mathrm{BCE}(S^\theta, \hat{S}^\theta). 
\end{align*}
Here $I_N^\theta$ and $\hat{I}^\theta_N$ are the predicted and ground-truth normal maps, $I_A^\theta$ and $\hat{I}^\theta_A$ are the predicted and ground-truth albedo maps, and $S^\theta$ and $\hat{S}^\theta$ are the predicted and ground-truth segmentation mask of the person. We empirically chose  $\lambda_a{=}0.5$, $\lambda_m{=}10$, $\lambda_\mathrm{VGG}{=}5$, and $\lambda_\mathrm{GAN}{=}0.1$.  We train the model at the resolution of 512$\times$512 for 230 epochs. 

\subsubsection{Personalization.} 
Given the RGB video of a novel subject, we optimize the latent features $Z$ and lighting environment $E$ of the video from scratch and only fine-tune AlbedoNet ($G_A$).  We keep $G_A$ fixed for the first $1000$ iterations and only optimize $Z$ and $E$. This allows optimization of the latent features $Z$ to be compatible with the pretrained $G_A$ and $G_N$. We then optimize $G_A$, $Z$, and $E$ jointly for a total of 15k iterations. As mentioned in Sec.~\ref{sec:rana}, we optimize the following objective 
\begin{align}
    \mathcal{L} =  \mathcal{L}_\mathrm{pixel} &+ \lambda_f\mathcal{L}_\mathrm{face} + 
    \lambda_m\mathcal{L}_\mathrm{mask} \\ &+ \lambda_\mathrm{VGG}\mathcal{L}_\mathrm{VGG} + \lambda_\mathrm{GAN}\mathcal{L}_\mathrm{GAN} \\  
    &+ \lambda_\mathrm{reg}^a\mathcal{L}_\mathrm{reg}^{\mathrm{albedo}} + \lambda_\mathrm{reg}^n\mathcal{L}_\mathrm{reg}^{\mathrm{normal}}. 
    \label{eq:rana_loss}
\end{align}
We chose $\lambda_f{=}1$, $\lambda_m{=}10$, $\lambda_\mathrm{VGG}{=}5$, $\lambda_\mathrm{GAN}{=}0.1$, $\lambda^a_\mathrm{reg}{=}0.5$ and $\lambda^n_\mathrm{reg}{=}0.25$,  For $\mathcal{L}_\mathrm{face}$, we project the nose keypoint of SMPL body model on to the image and crop a $100{\times100}$ patch around the face to compute the loss. All other losses are calculated on the $512{\times}512$ generated images as mentioned above.

\subsection{Relighting Human Dataset} 
\label{sec:dataset}
\input{figures/fig_example_subjects_RH.tex}
\input{figures/fig_train_test_protocol_a.tex}
\input{figures/fig_train_test_protocol_b.tex}

We provide more details about our proposed Relighting Humans dataset. The dataset consists of $49$ subjects with 26 males and 23 female characters. The characters come in all appearances, including long hair, loose clothing, jackets, hats, head scarves, etc. Some example characters can be seen in Fig.~\ref{fig:example_subjects_rh}. Moreover, we also provide some examples of training and testing sequences for protocol-a and protocol-b in Fig.~\ref{fig:train_test_protocol_a} and Fig.~\ref{fig:train_test_protocol_b}, respectively. 

\subsection{Qualitative comparison with SMPL+D baseline}
\label{sec:rana_vs_smpld}
As reported in Tab.~\ref{tab:baselines_and_sota}, SMPL+D baseline yields better quantitative results than RANA even though it provides overly smoothed texture details due to TextureNet confusing shading and texture during albedo texture map estimation. On the other hand, RANA recovers significantly better texture details, but sometimes lighting can still leak into albedo texture, especially for very bright or complex lighting environments. We found that the used evaluation metrics penalize color differences more than the texture details as we show in Fig.~\ref{fig:rana_vs_smpld}. In any case, RANA provides significantly better results for final image reconstruction as shown in Tab.~\ref{tab:baselines_and_sota}. 

\input{figures/fig_rana_vs_smpld.tex}

%% file: figures/fig_example_subjects_RH.tex
\begin{figure*}
\centering
\scalebox{0.97}{
\small
\setlength{\tabcolsep}{1pt}
\renewcommand{\arraystretch}{0.4}
  \begin{tabular}{cccc}
     Image (Path Traced) & Albedo Image & Nomral Map & Segmentation Mask  \\
    \includegraphics[trim={2cm 0cm 2cm 0cm},clip,width=0.25\textwidth]{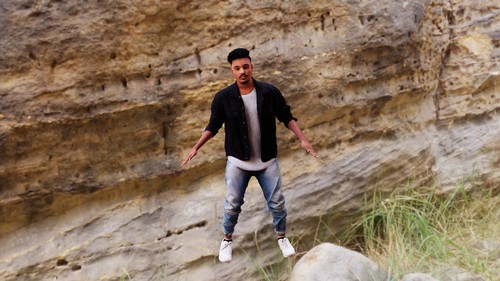} & 
    \includegraphics[trim={2cm 0cm 2cm 0cm},clip,width=0.25\textwidth]{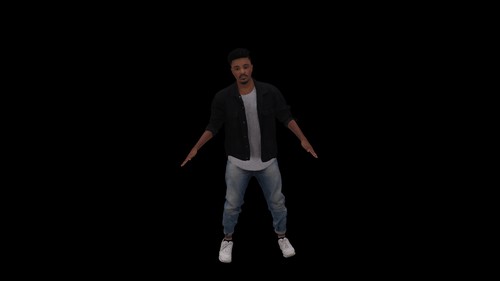} &
    \includegraphics[trim={2cm 0cm 2cm 0cm},clip,width=0.25\textwidth]{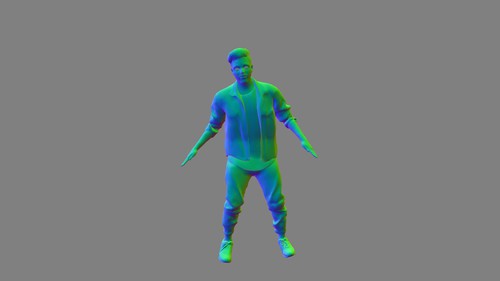} &
    \includegraphics[trim={2cm 0cm 2cm 0cm},clip,width=0.25\textwidth]{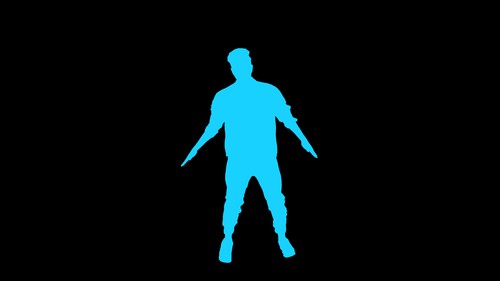} \\
    \includegraphics[trim={2cm 0cm 2cm 0cm},clip,width=0.25\textwidth]{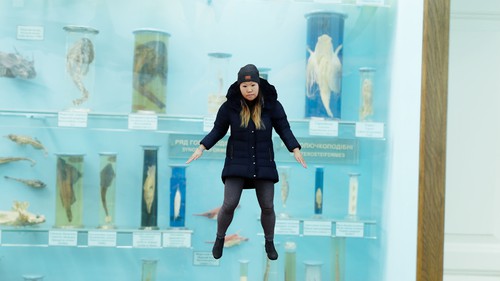} & 
    \includegraphics[trim={2cm 0cm 2cm 0cm},clip,width=0.25\textwidth]{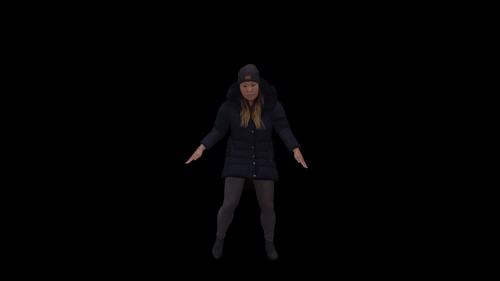} &
    \includegraphics[trim={2cm 0cm 2cm 0cm},clip,width=0.25\textwidth]{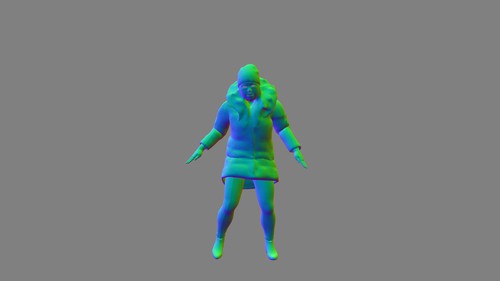} &
    \includegraphics[trim={2cm 0cm 2cm 0cm},clip,width=0.25\textwidth]{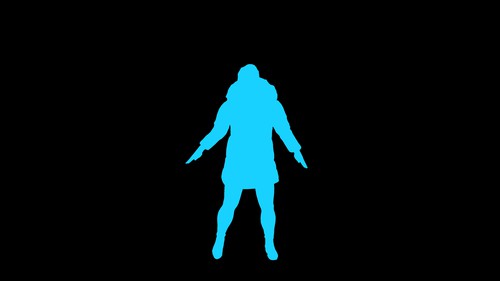} \\
    \includegraphics[trim={2cm 0cm 2cm 0cm},clip,width=0.25\textwidth]{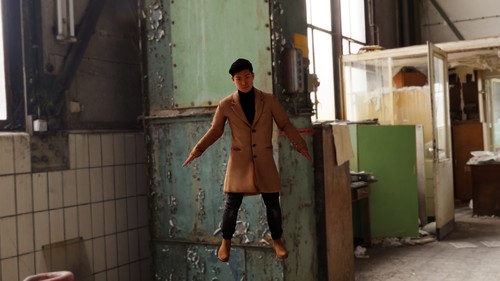} & 
    \includegraphics[trim={2cm 0cm 2cm 0cm},clip,width=0.25\textwidth]{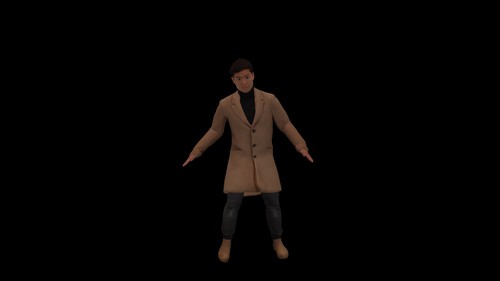} &
    \includegraphics[trim={2cm 0cm 2cm 0cm},clip,width=0.25\textwidth]{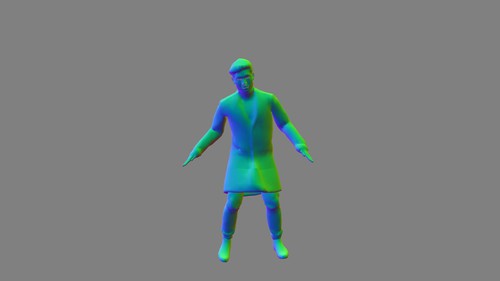} &
    \includegraphics[trim={2cm 0cm 2cm 0cm},clip,width=0.25\textwidth]{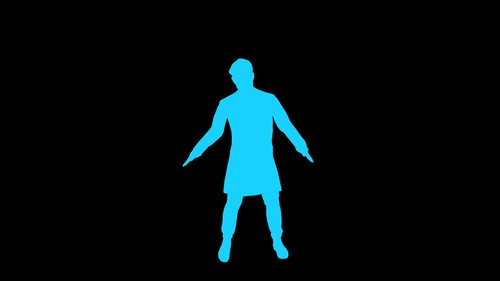} \\
    \includegraphics[trim={2cm 0cm 2cm 0cm},clip,width=0.25\textwidth]{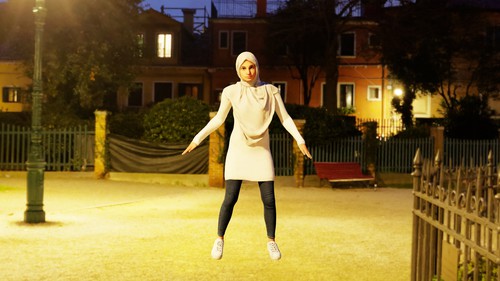} & 
    \includegraphics[trim={2cm 0cm 2cm 0cm},clip,width=0.25\textwidth]{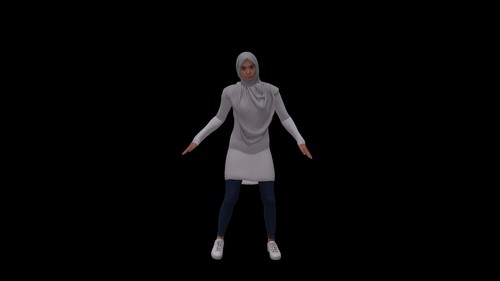} &
    \includegraphics[trim={2cm 0cm 2cm 0cm},clip,width=0.25\textwidth]{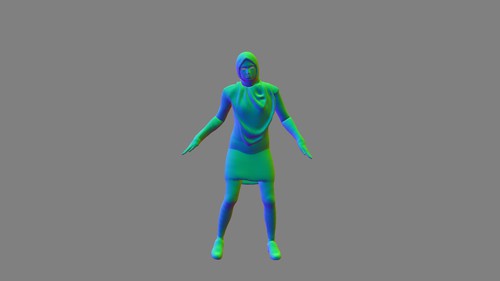} &
    \includegraphics[trim={2cm 0cm 2cm 0cm},clip,width=0.25\textwidth]{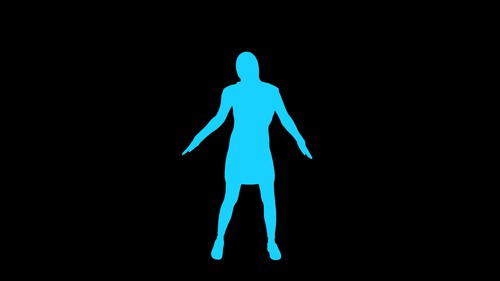} \\
    \includegraphics[trim={2cm 0cm 2cm 0cm},clip,width=0.25\textwidth]{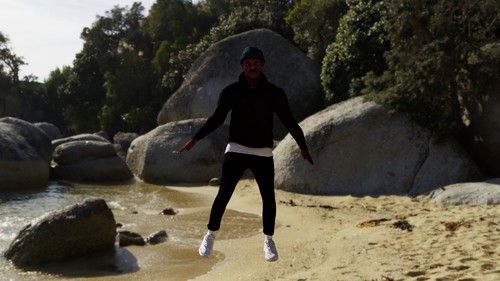} & 
    \includegraphics[trim={2cm 0cm 2cm 0cm},clip,width=0.25\textwidth]{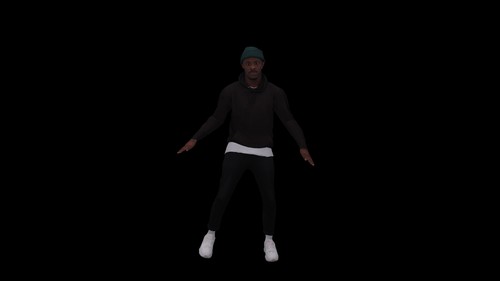} &
    \includegraphics[trim={2cm 0cm 2cm 0cm},clip,width=0.25\textwidth]{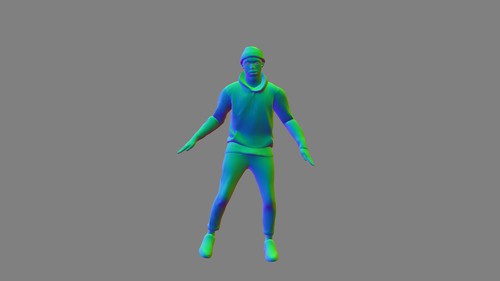} &
    \includegraphics[trim={2cm 0cm 2cm 0cm},clip,width=0.25\textwidth]{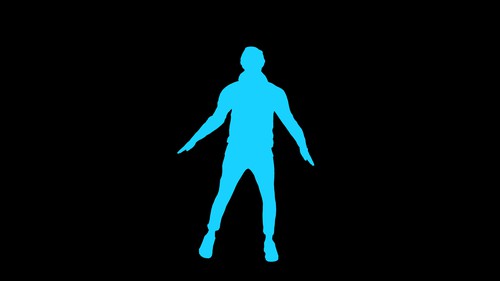} \\
    \includegraphics[trim={2cm 0cm 2cm 0cm},clip,width=0.25\textwidth]{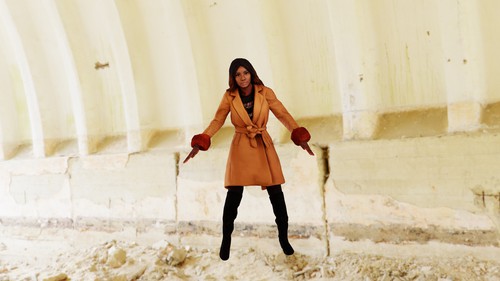} & 
    \includegraphics[trim={2cm 0cm 2cm 0cm},clip,width=0.25\textwidth]{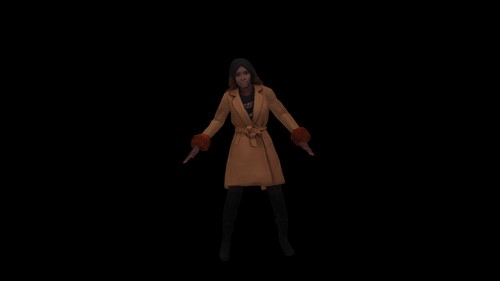} &
    \includegraphics[trim={2cm 0cm 2cm 0cm},clip,width=0.25\textwidth]{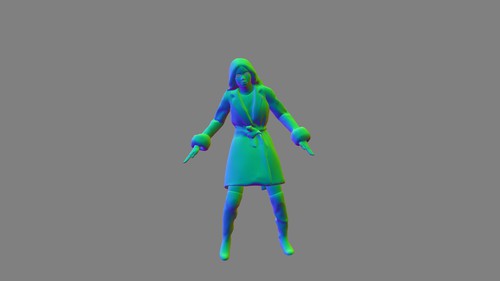} &
    \includegraphics[trim={2cm 0cm 2cm 0cm},clip,width=0.25\textwidth]{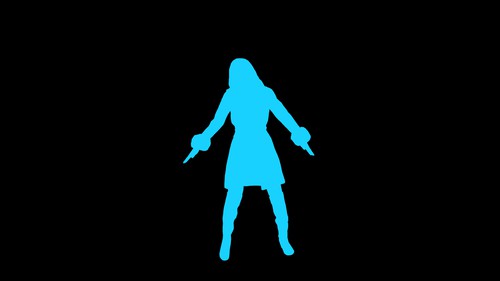} \\
    \includegraphics[trim={2cm 0cm 2cm 0cm},clip,width=0.25\textwidth]{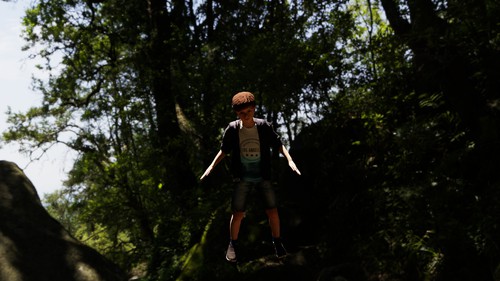} & 
    \includegraphics[trim={2cm 0cm 2cm 0cm},clip,width=0.25\textwidth]{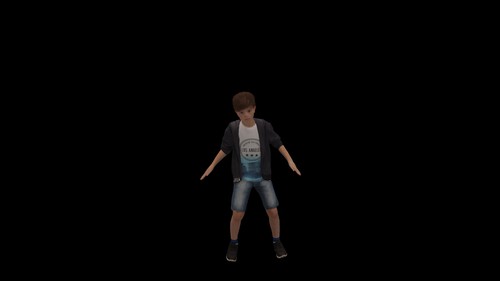} &
    \includegraphics[trim={2cm 0cm 2cm 0cm},clip,width=0.25\textwidth]{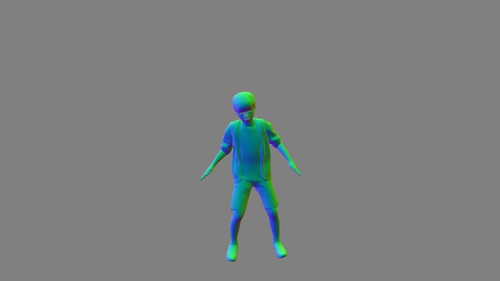} &
    \includegraphics[trim={2cm 0cm 2cm 0cm},clip,width=0.25\textwidth]{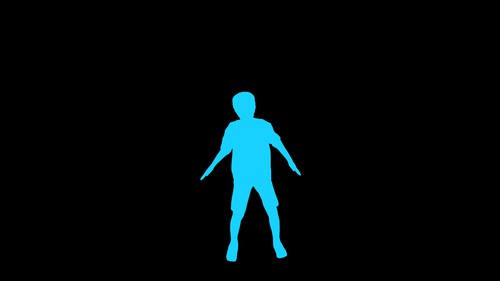} \\
  \end{tabular}
}
\caption{Example subjects from our proposed Relighting Humans dataset. We provide the ground-truth albedo map, normal map and segmentation masks as the ground truths.}
\label{fig:example_subjects_rh} 
\end{figure*}
 

%% file: figures/fig_train_test_protocol_a.tex
\begin{figure*}
\centering
\scalebox{1}{
    \setlength{\tabcolsep}{1pt}
  \begin{tabular}{c|ccc}
    \multicolumn{1}{c|}{Training Frames} & \multicolumn{3}{c}{Testing Frames} \\
    \includegraphics[trim={2cm 0cm 2cm 0cm},clip,width=0.24\textwidth]{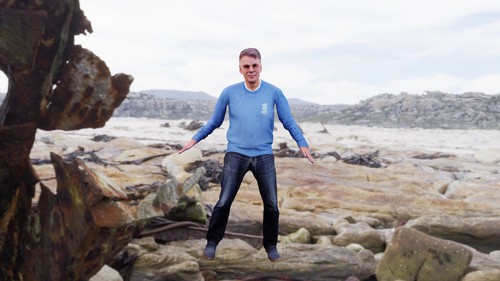} &
    \includegraphics[trim={2cm 0cm 2cm 0cm},clip,width=0.24\textwidth]{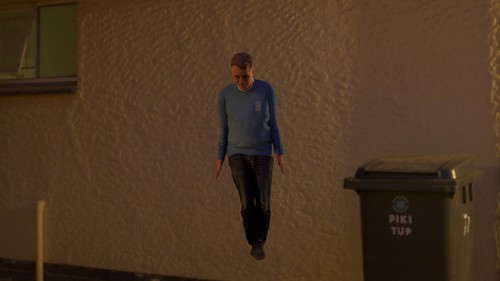} & 
    \includegraphics[trim={2cm 0cm 2cm 0cm},clip,width=0.24\textwidth]{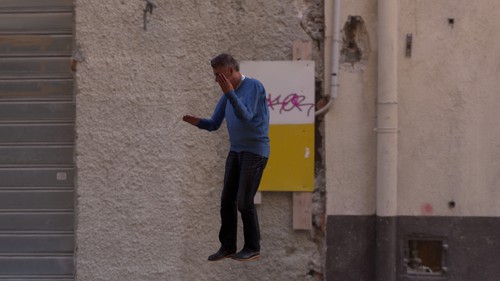} &  
    \includegraphics[trim={2cm 0cm 2cm 0cm},clip,width=0.24\textwidth]{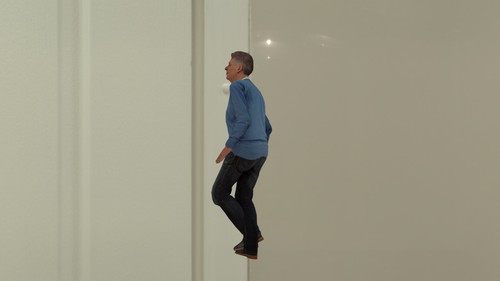} 
    \\
    \includegraphics[trim={2cm 0cm 2cm 0cm},clip,width=0.24\textwidth]{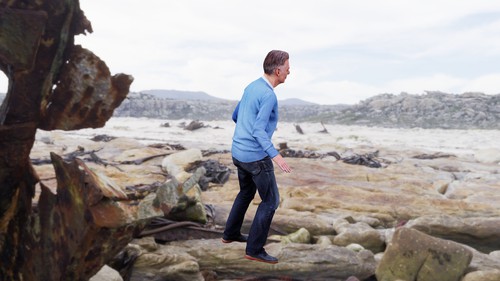} &
    \includegraphics[trim={2cm 0cm 2cm 0cm},clip,width=0.24\textwidth]{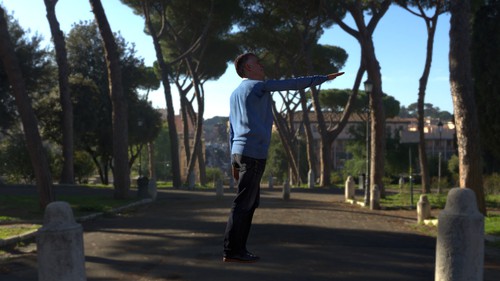} & 
    \includegraphics[trim={2cm 0cm 2cm 0cm},clip,width=0.24\textwidth]{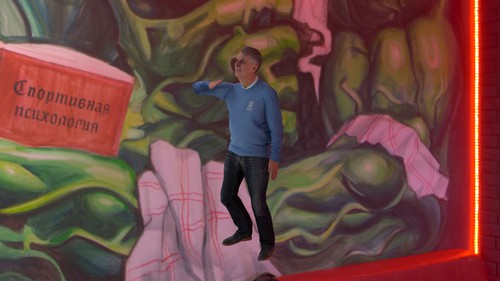} &  
    \includegraphics[trim={2cm 0cm 2cm 0cm},clip, width=0.24\textwidth]{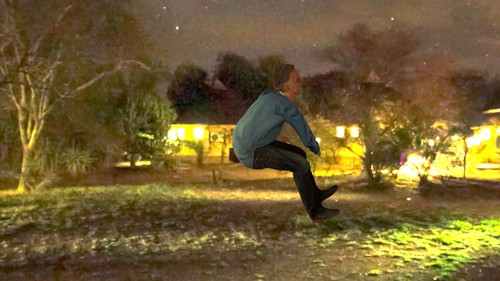} 
    \\
    \includegraphics[trim={2cm 0cm 2cm 0cm},clip,width=0.24\textwidth]{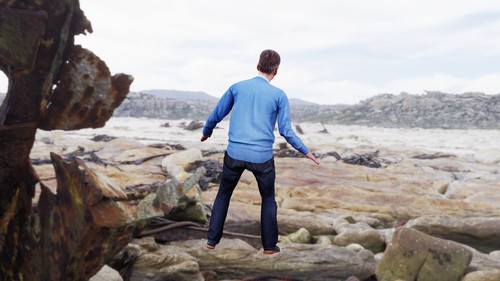} &
    \includegraphics[trim={2cm 0cm 2cm 0cm},clip,width=0.24\textwidth]{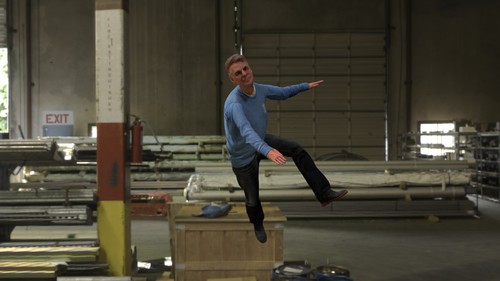} & 
    \includegraphics[trim={2cm 0cm 2cm 0cm},clip,width=0.24\textwidth]{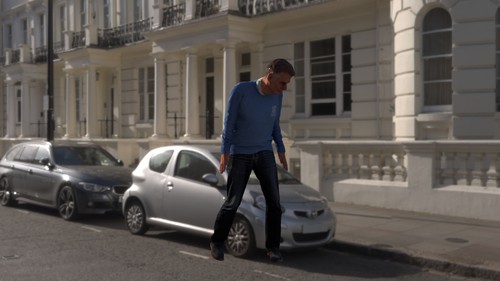} &  
    \includegraphics[trim={2cm 0cm 2cm 0cm},clip, width=0.24\textwidth]{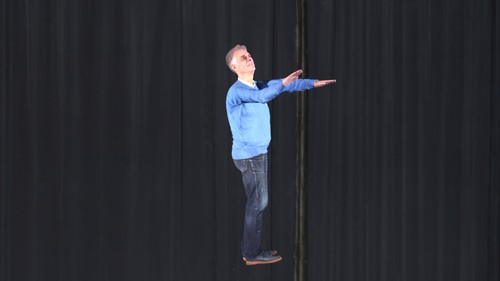} 
    \\
    \midrule
    \midrule
    \includegraphics[trim={2cm 0cm 2cm 0cm},clip,width=0.24\textwidth]{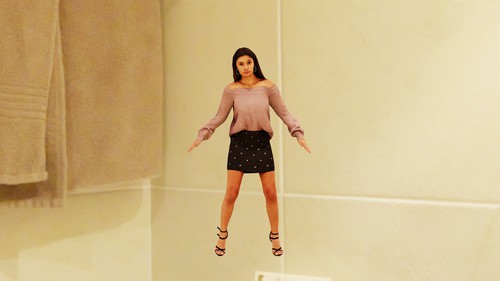} &
    \includegraphics[trim={2cm 0cm 2cm 0cm},clip,width=0.24\textwidth]{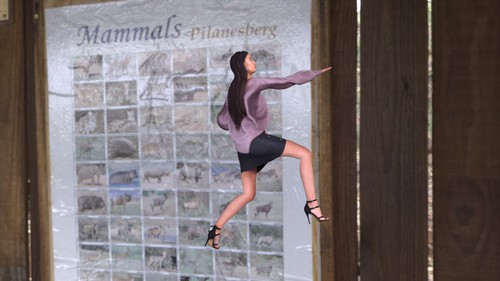} & 
    \includegraphics[trim={2cm 0cm 2cm 0cm},clip,width=0.24\textwidth]{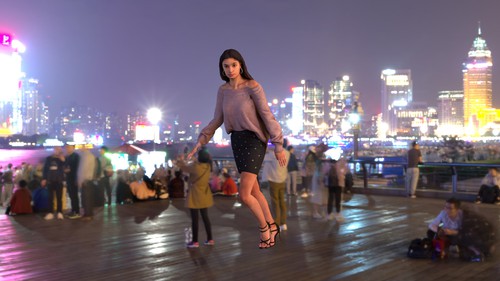} &  
    \includegraphics[trim={2cm 0cm 2cm 0cm},clip, width=0.24\textwidth]{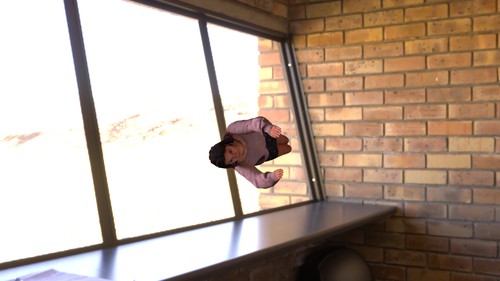} 
    \\
    \includegraphics[trim={2cm 0cm 2cm 0cm},clip,width=0.24\textwidth]{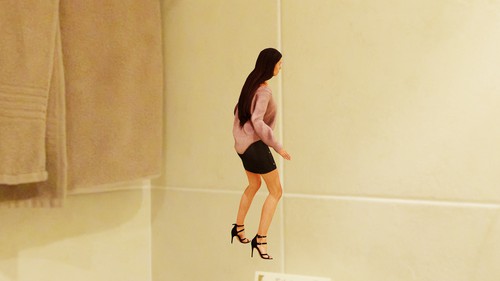} &
    \includegraphics[trim={2cm 0cm 2cm 0cm},clip,width=0.24\textwidth] {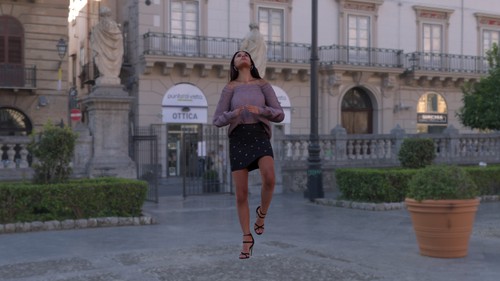} & 
    \includegraphics[trim={2cm 0cm 2cm 0cm},clip,width=0.24\textwidth]{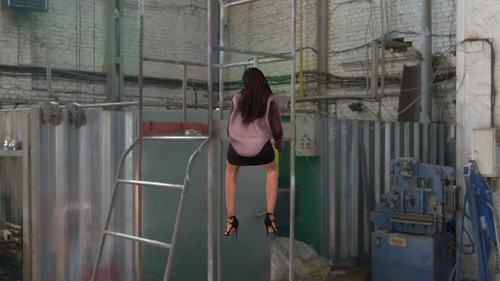} &  
    \includegraphics[trim={2cm 0cm 2cm 0cm},clip, width=0.24\textwidth]{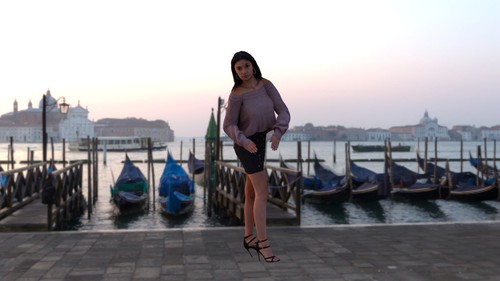} 
    \\
    \includegraphics[trim={2cm 0cm 2cm 0cm},clip,width=0.24\textwidth]{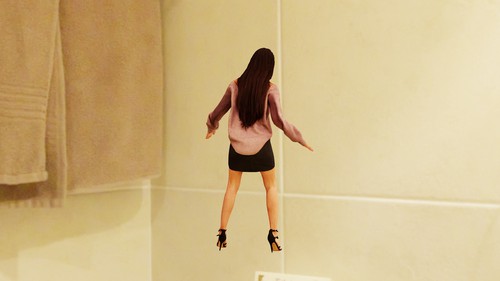} &
    \includegraphics[trim={2cm 0cm 2cm 0cm},clip,width=0.24\textwidth]{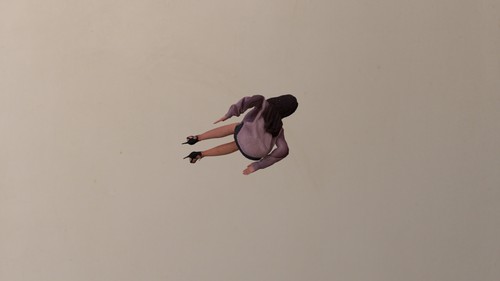} & 
    \includegraphics[trim={2cm 0cm 2cm 0cm},clip,width=0.24\textwidth]{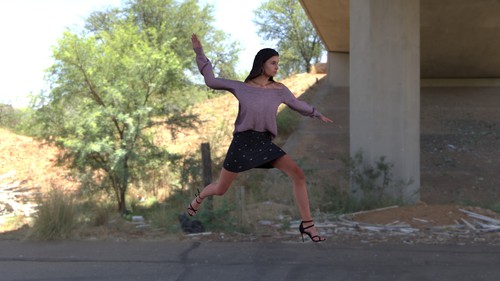} &  
    \includegraphics[trim={2cm 0cm 2cm 0cm},clip, width=0.24\textwidth]{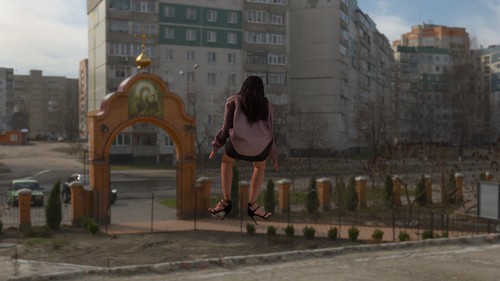} 
    \\
  \end{tabular}
}
\vspace{-4mm}
\caption{Example training (column-1) and testing (column 2-4) samples for \textbf{Protocol-a} of the proposed Relighting Humans dataset. The training frames are generated in an A-pose with the same lighting, while each of the test frames has a random body pose and lighting environment. }
\label{fig:train_test_protocol_a}
\end{figure*}

%% file: figures/fig_train_test_protocol_b.tex
\begin{figure*}
\centering
\scalebox{1}{
    \setlength{\tabcolsep}{1pt}
  \begin{tabular}{ccccc}
    \multicolumn{1}{c}{\rotatebox{90}{~~~~~~~~~~~Train}} & 
    \includegraphics[trim={1cm 0cm 1cm 0cm},clip,width=0.24\textwidth]{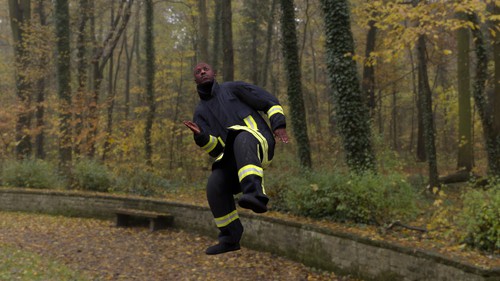} &
    \includegraphics[trim={1cm 0cm 1cm 0cm},clip,width=0.24\textwidth]{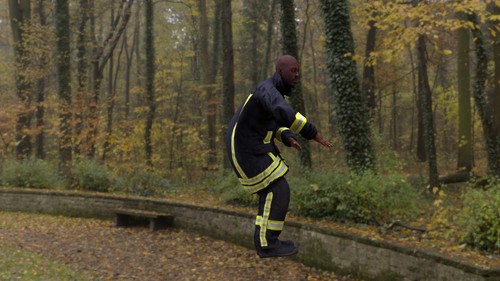} &  
    \includegraphics[trim={1cm 0cm 1cm 0cm},clip,width=0.24\textwidth]{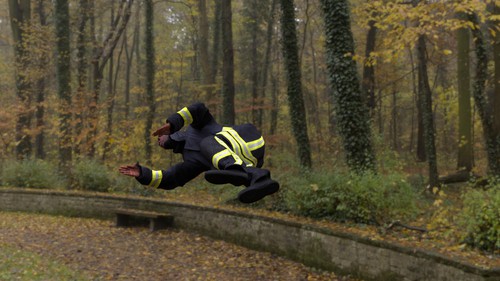} &
    \includegraphics[trim={1cm 0cm 1cm 0cm},clip,width=0.24\textwidth]{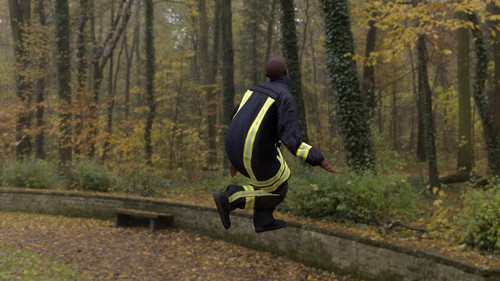} \\
    \rotatebox{90}{~~~~~~~~~~~~Test} & 
    \includegraphics[trim={1cm 0cm 1cm 0cm},clip,width=0.24\textwidth]{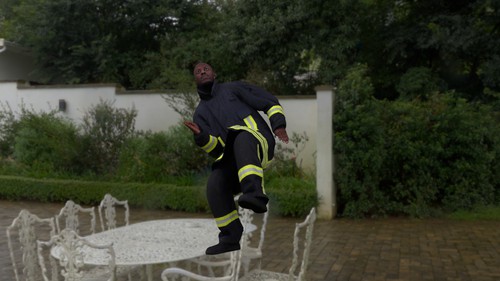} & 
    \includegraphics[trim={1cm 0cm 1cm 0cm},clip, width=0.24\textwidth]{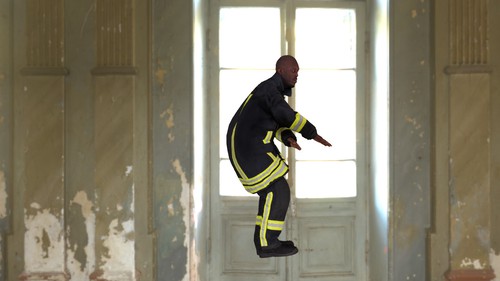} & 
    \includegraphics[trim={1cm 0cm 1cm 0cm},clip,width=0.24\textwidth]{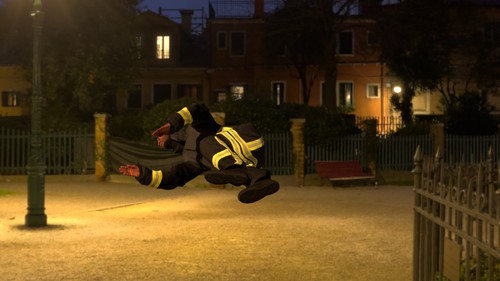} & 
    \includegraphics[trim={1cm 0cm 1cm 0cm},clip, width=0.24\textwidth]{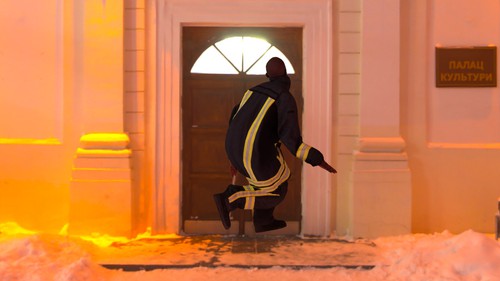} \\ 
    \midrule
    \midrule
    \multicolumn{1}{c}{\rotatebox{90}{~~~~~~~~~~~Train}} & 
    \includegraphics[trim={1cm 0cm 1cm 0cm},clip,width=0.24\textwidth]{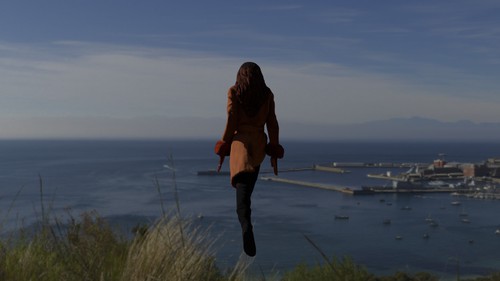} &
    \includegraphics[trim={1cm 0cm 1cm 0cm},clip,width=0.24\textwidth]{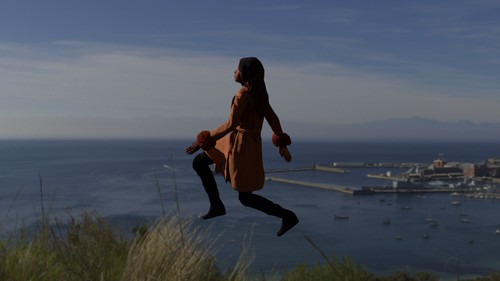} &  
    \includegraphics[trim={1cm 0cm 1cm 0cm},clip,width=0.24\textwidth]{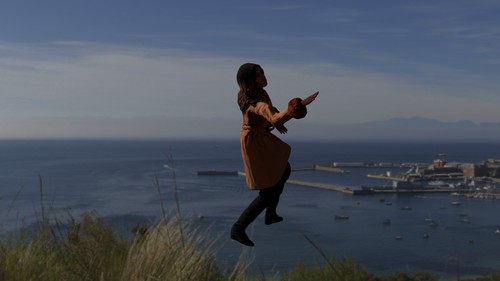} &
    \includegraphics[trim={1cm 0cm 1cm 0cm},clip,width=0.24\textwidth]{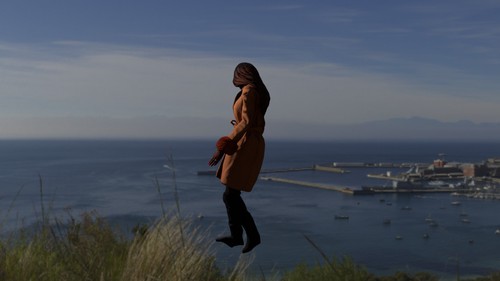} \\
    \rotatebox{90}{~~~~~~~~~~~~Test} & 
    \includegraphics[trim={1cm 0cm 1cm 0cm},clip,width=0.24\textwidth]{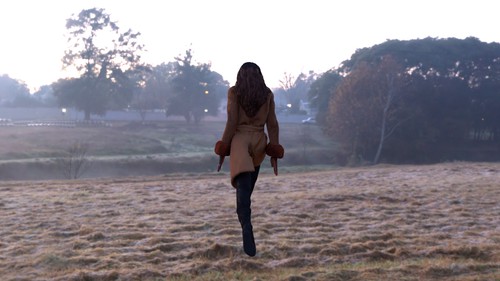} & 
    \includegraphics[trim={1cm 0cm 1cm 0cm},clip, width=0.24\textwidth]{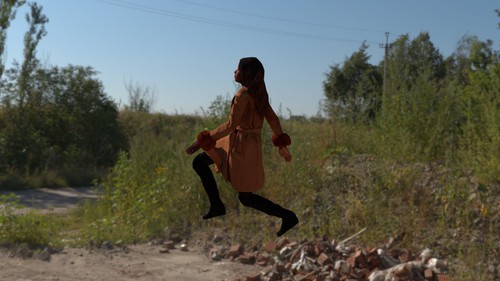} & 
    \includegraphics[trim={1cm 0cm 1cm 0cm},clip,width=0.24\textwidth]{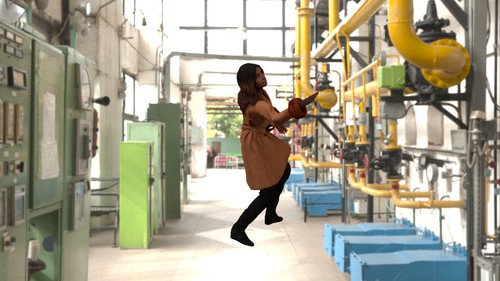} & 
    \includegraphics[trim={1cm 0cm 1cm 0cm},clip, width=0.24\textwidth]{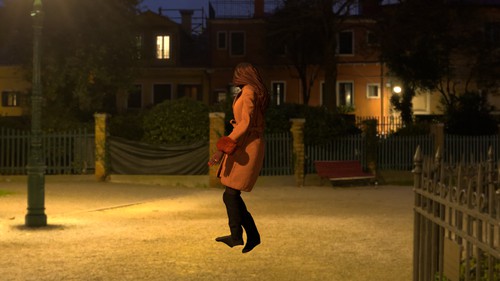} \\    
    \midrule
    \midrule
    \multicolumn{1}{c}{\rotatebox{90}{~~~~~~~~~~~Train}} & 
    \includegraphics[trim={1cm 0cm 1cm 0cm},clip,width=0.24\textwidth]{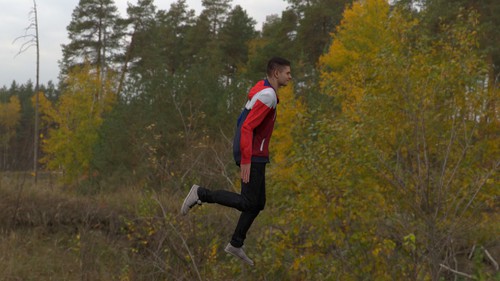} &
    \includegraphics[trim={1cm 0cm 1cm 0cm},clip,width=0.24\textwidth]{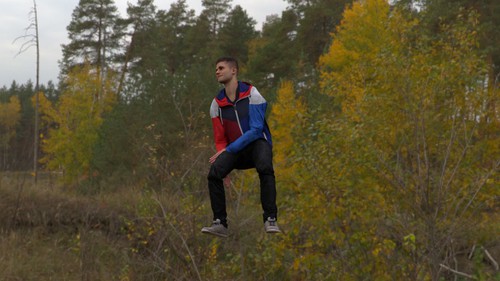} &  
    \includegraphics[trim={1cm 0cm 1cm 0cm},clip,width=0.24\textwidth]{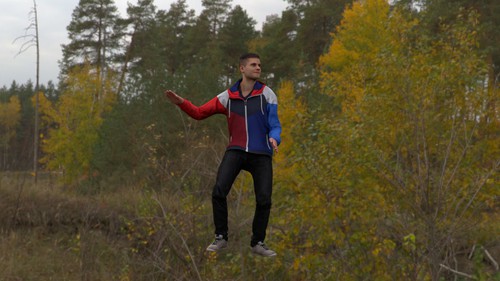} &
    \includegraphics[trim={1cm 0cm 1cm 0cm},clip,width=0.24\textwidth]{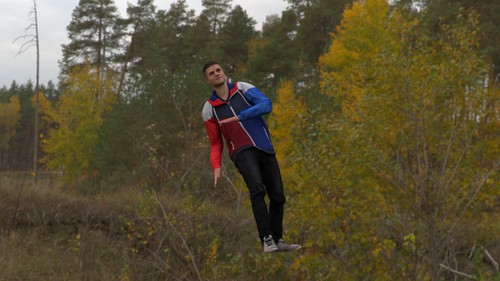} \\
    \rotatebox{90}{~~~~~~~~~~~~Test} & 
    \includegraphics[trim={1cm 0cm 1cm 0cm},clip,width=0.24\textwidth]{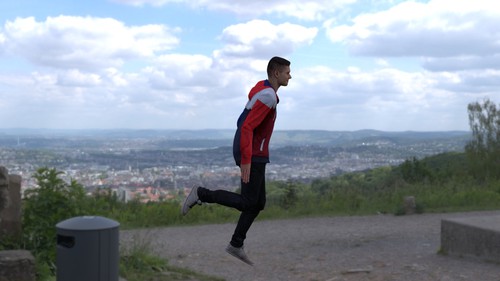} & 
    \includegraphics[trim={1cm 0cm 1cm 0cm},clip, width=0.24\textwidth]{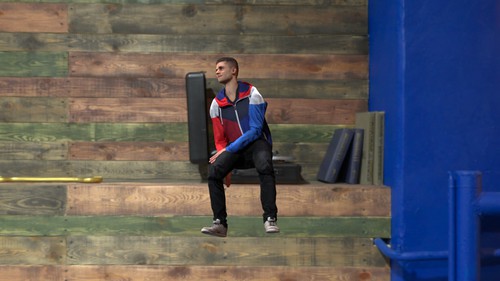} & 
    \includegraphics[trim={1cm 0cm 1cm 0cm},clip,width=0.24\textwidth]{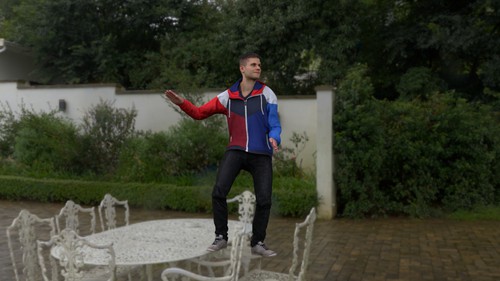} & 
    \includegraphics[trim={1cm 0cm 1cm 0cm},clip, width=0.24\textwidth]{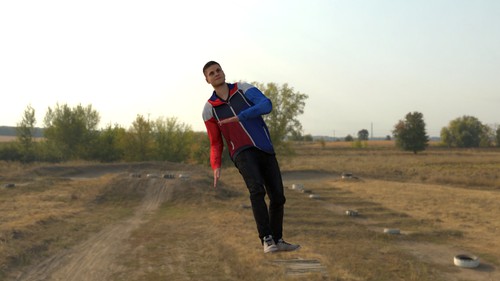} \\

  \end{tabular}
}
\vspace{-4mm}
\caption{Example training (rows 1,3,5) and testing (rows 2,4,6) samples for \textbf{Protocol-b} of the proposed Relighting Humans dataset. The training frames are generated with fixed lighting and random body poses. The testing frames have exactly the same body poses as training frames but with different lighting environments.\vspace{-5mm}}
\label{fig:train_test_protocol_b}
\end{figure*}

%% file: figures/fig_rana_vs_smpld.tex
\begin{figure*}
\vspace{-7mm}
\centering
\scalebox{1}{
    \setlength{\tabcolsep}{1pt}
  \begin{tabular}{cc}
    ~~~~~~~GT Albedo Map ~~~~~~~~~~ RANA Albedo Map~~~~~~~~ SMPL Albedo Map~~~~~ & Reference Image \\
    \includegraphics[height=0.15\textheight]{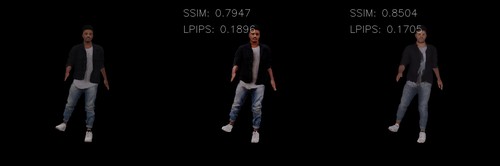} & \includegraphics[trim={3cm 0cm 3cm 0cm},clip,height=0.15\textheight]{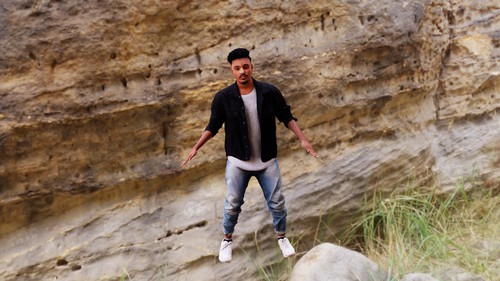}\\
    \includegraphics[height=0.15\textheight]{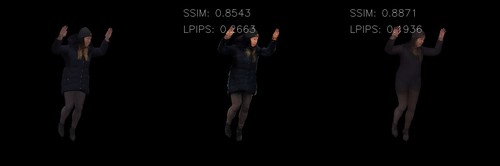} & \includegraphics[trim={3cm 0cm 3cm 0cm},clip,height=0.15\textheight]{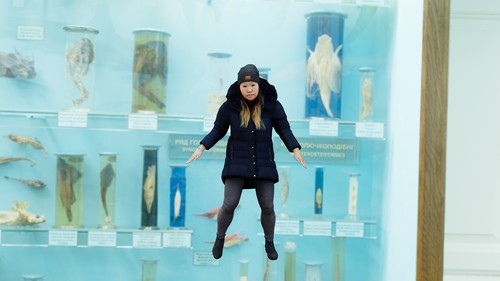} \\
    \includegraphics[height=0.15\textheight]{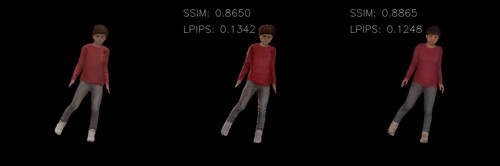} & \includegraphics[trim={3cm 0cm 3cm 0cm},clip,height=0.15\textheight]{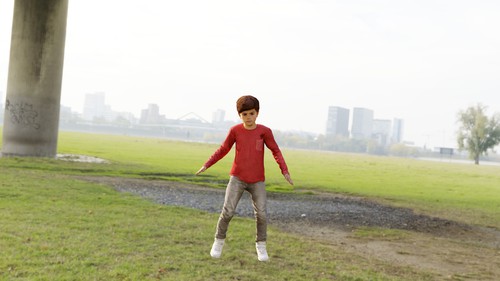} \\
    \includegraphics[height=0.15\textheight]{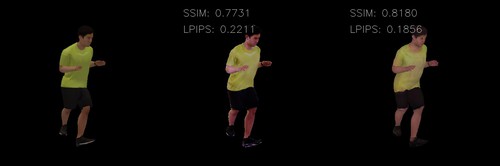} & \includegraphics[trim={3cm 0cm 3cm 0cm},clip,height=0.15\textheight]{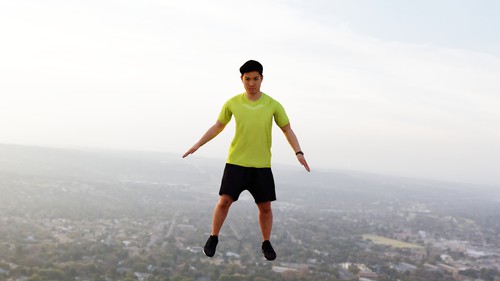} \\
    \includegraphics[height=0.15\textheight]{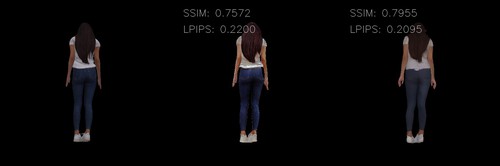} & \includegraphics[trim={3cm 0cm 3cm 0cm},clip,height=0.15\textheight]{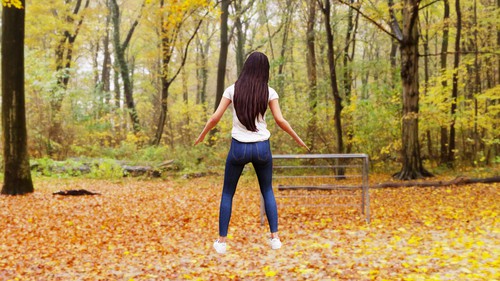} \\
    \includegraphics[height=0.15\textheight]{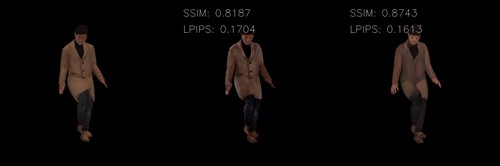} & \includegraphics[trim={3cm 0cm 3cm 0cm},clip,height=0.15\textheight]{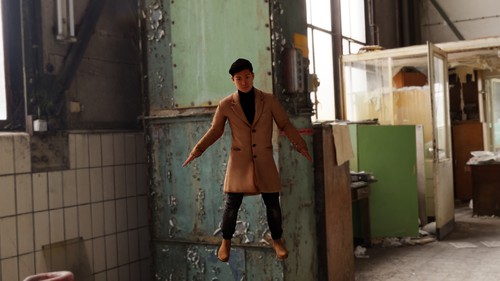} \\

  \end{tabular}
}
\vspace{-4mm}
\caption{Qualitative comparison with SMPL+D baseline for albedo map reconstruction. We show the ground-truth albedo maps (column 1), reconstructed albedo maps by RANA (column 2), and the reconstructed albedo map by SMPL+D baseline (column 3). We also show a reference training frame (column 4) which is used to create the avatar. We overlay the SSIM and LPIPS scores on the reconstructed albedo maps. SMPL+D yields better quantitative metrics even though it generates overly smooth albedo maps. In contrast, RANA provides significantly better texture details but sometimes the light information still leaks into the albedo textures. The SSIM and LPIPS metrics seem to penalize more for color difference than missing texture details. \vspace{-5mm}}
\label{fig:rana_vs_smpld}
\end{figure*}

%% file: main.bbl
\begin{thebibliography}{10}\itemsep=-1pt

\bibitem{polyhaven}
{Poly Haven}, 2020.
\newblock \url{https://hdrihaven.com}.

\bibitem{renderpeople}
{R}ender {P}eople, 2020.
\newblock \url{https://renderpeople.com/3d-people}.

\bibitem{cmu_mocap}
Carnegie {M}ellon {U}niversity {G}raphics {L}ab: Motion {C}apture {D}atabase.
\newblock \url{http://mocap.cs.cmu.edu}, 2014.

\bibitem{alldieck2018detailed}
Thiemo Alldieck, Marcus Magnor, Weipeng Xu, Christian Theobalt, and Gerard
  Pons-Moll.
\newblock Detailed human avatars from monocular video.
\newblock In {\em 3DV}, 2018.

\bibitem{Alldieck_2018_CVPR}
Thiemo Alldieck, Marcus Magnor, Weipeng Xu, Christian Theobalt, and Gerard
  Pons-Moll.
\newblock Video based reconstruction of 3d people models.
\newblock In {\em CVPR}, June 2018.

\bibitem{alldieck2022phorhum}
Thiemo Alldieck, Mihai Zanfir, and Cristian Sminchisescu.
\newblock Photorealistic monocular 3d reconstruction of humans wearing
  clothing.
\newblock In {\em CVPR}, 2022.

\bibitem{balakrishnan2018synthesizing}
Guha Balakrishnan, Amy Zhao, Adrian~V Dalca, Fredo Durand, and John Guttag.
\newblock Synthesizing images of humans in unseen poses.
\newblock In {\em CVPR}, 2018.

\bibitem{barron2020shape}
Jonathan~T. Barron and Jitendra Malik.
\newblock Shape, illumination, and reflectance from shading.
\newblock In {\em TPAMI}, 2020.

\bibitem{bogo2015detailed}
Federica Bogo, Michael~J Black, Matthew Loper, and Javier Romero.
\newblock Detailed full-body reconstructions of moving people from monocular
  rgb-d sequences.
\newblock In {\em ICCV}, 2015.

\bibitem{carranza2003free}
Joel Carranza, Christian Theobalt, Marcus~A Magnor, and Hans-Peter Seidel.
\newblock Free-viewpoint video of human actors.
\newblock {\em ToG}, 2003.

\bibitem{chan2019everybody}
Caroline Chan, Shiry Ginosar, Tinghui Zhou, and Alexei~A Efros.
\newblock Everybody dance now.
\newblock In {\em ICCV}, 2019.

\bibitem{check2017deeplab}
Liang-Chieh Chen, George Papandreou, Florian Schroff, and Hartwig Adam.
\newblock Rethinking atrous convolution for semantic image segmentation.
\newblock In {\em CVPR}, 2017.

\bibitem{chen2022relighting}
Zhaoxi Chen and Ziwei Liu.
\newblock Relighting4d: Neural relightable human from videos.
\newblock In {\em ECCV}, 2022.

\bibitem{debevec2000acquiring}
Paul Debevec, Tim Hawkins, Chris Tchou, Haarm-Pieter Duiker, Westley Sarokin,
  and Mark Sagar.
\newblock Acquiring the reflectance field of a human face.
\newblock In {\em Annual conference on Computer Graphics and Interactive
  Techniques}, 2000.

\bibitem{ding2022dists}
Keyan Ding, Kede Ma, Shiqi Wang, and Eero~P. Simoncelli.
\newblock {Image Quality Assessment: Unifying Structure and Texture
  Similarity}.
\newblock {\em TPAMI}, 2022.

\bibitem{esser2018variational}
Patrick Esser, Ekaterina Sutter, , and Bjorn Ommer.
\newblock A variational {U-Net} for conditional appearance and shape
  generation.
\newblock In {\em CVPR}, 2018.

\bibitem{goodfellow2014gan}
Ian Goodfellow, Jean Pouget-Abadie, Mehdi Mirza, Bing Xu, David Warde-Farley,
  Sherjil Ozair, Aaron Courville, and Yoshua Bengio.
\newblock Generative adversarial nets.
\newblock In {\em NeurIPS}, 2014.

\bibitem{grigorev2021stylepeople}
Artur Grigorev, Karim Iskakov, Anastasia Ianina, Renat Bashirov, Ilya
  Zakharkin, Alexander Vakhitov, and Victor Lempitsky.
\newblock Stylepeople: A generative model of fullbody human avatars.
\newblock In {\em CVPR}, 2021.

\bibitem{guo2019relightables}
Kaiwen Guo, Peter Lincoln, Philip Davidson, Jay Busch, Xueming Yu, Matt Whalen,
  Geoff Harvey, Sergio Orts-Escolano, Rohit Pandey, Jason Dourgarian, et~al.
\newblock The relightables: volumetric performance capture of humans with
  realistic relighting.
\newblock In {\em TOG}, 2019.

\bibitem{He2021ARCHAC}
Tong He, Yuanlu Xu, Shunsuke Saito, Stefano Soatto, and Tony Tung.
\newblock Arch++: Animation-ready clothed human reconstruction revisited.
\newblock In {\em ICCV}, 2021.

\bibitem{huang2021few}
Zhichao Huang, Xintong Han, Jia Xu, and Tong Zhang.
\newblock Few-shot human motion transfer by personalized geometry and texture
  modeling.
\newblock In {\em CVPR}, 2021.

\bibitem{huang2020arch}
Zeng Huang, Yuanlu Xu, Christoph Lassner, Hao Li, and Tony Tung.
\newblock {ARCH}: Animatable reconstruction of clothed humans.
\newblock In {\em CVPR}, 2020.

\bibitem{iqbal2021kama}
Umar Iqbal, Kevin Xie, Yunrong Guo, Jan Kautz, and Pavlo Molchanov.
\newblock Kama: 3d keypoint aware body mesh articulation.
\newblock In {\em 3DV}, 2021.

\bibitem{isola2017image}
Phillip Isola, Jun-Yan Zhu, Tinghui Zhou, and Alexei~A Efros.
\newblock Image-to-image translation with conditional adversarial networks.
\newblock In {\em CVPR}, 2017.

\bibitem{ji2022geometry}
Chaonan Ji, Tao Yu, Kaiwen Guo, Jingxin Liu, and Yebin Liu.
\newblock Geometry-aware single-image full-body human relighting.
\newblock In {\em ECCV}, 2022.

\bibitem{jiang2022neuman}
Wei Jiang, Kwang~Moo Yi, Golnoosh Samei, Oncel Tuzel, and Anurag Ranjan.
\newblock Neuman: Neural human radiance field from a single video.
\newblock In {\em ECCV}, 2022.

\bibitem{kanamori_relight2018}
Yoshihiro Kanamori and Yuki Endo.
\newblock Relighting humans: Occlusion-aware inverse rendering for full-body
  human images.
\newblock In {\em SIGGRAPH}, 2018.

\bibitem{kanamori2018relighing}
Yoshihiro Kanamori and Yuki Endo.
\newblock Relighting humans: Occlusion-aware inverse rendering for full-body
  human images.
\newblock In {\em SIGGRAPH Asia}, 2018.

\bibitem{Lagunas2021humanrelighting}
Manuel Lagunas, Xin Sun, Jimei Yang, Ruben Villegas, Jianming Zhang, Zhixin
  Shu, Belen Masia, and Diego Gutierrez.
\newblock Single-image full-body human relighting.
\newblock In {\em Eurographics Symposium on Rendering (EGSR)}, 2021.

\bibitem{lazova3dv2019}
Verica Lazova, Eldar Insafutdinov, and Gerard Pons-Moll.
\newblock 360-degree textures of people in clothing from a single image.
\newblock In {\em 3DV}, 2019.

\bibitem{li2013capturing}
Guannan Li, Chenglei Wu, Carsten Stoll, Yebin Liu, Kiran Varanasi, Qionghai
  Dai, and Christian Theobalt.
\newblock Capturing relightable human performances under general uncontrolled
  illumination.
\newblock In {\em Computer Graphics Forum}. Wiley Online Library, 2013.

\bibitem{liu2019softras}
Shichen Liu, Tianye Li, Weikai Chen, and Hao Li.
\newblock Soft rasterizer: A differentiable renderer for image-based 3d
  reasoning.
\newblock In {\em ICCV}, 2019.

\bibitem{loper2015smpl}
Matthew Loper, Naureen Mahmood, Javier Romero, Gerard Pons-Moll, and Michael~J.
  Black.
\newblock {SMPL}: A skinned multi-person linear model.
\newblock {\em SIGGRAPH ASIA}, 2015.

\bibitem{meka2022deeprelightable}
Abhimitra Meka, Rohit Pandey, Christian Haene, Sergio Orts-Escolano, Peter
  Barnum, Philip Davidson, Daniel Erickson, Yinda Zhang, Jonathan Taylor,
  Sofien Bouaziz, Chloe Legendre, Wan-Chun Ma, Ryan Overbeck, Thabo Beeler,
  Paul Debevec, Shahram Izadi, Christian Theobalt, Christoph Rhemann, and Sean
  Fanello.
\newblock Deep relightable textures - volumetric performance capture with
  neural rendering.
\newblock {\em SIGGRAPH}, 39(6), 2020.

\bibitem{mildenhall2020nerf}
Ben Mildenhall, Pratul~P. Srinivasan, Matthew Tancik, Jonathan~T. Barron, Ravi
  Ramamoorthi, and Ren Ng.
\newblock {NeRF}: {R}epresenting scenes as neural radiance fields for view
  synthesis.
\newblock In {\em ECCV}, 2020.

\bibitem{pandey2021total}
Rohit Pandey, Sergio~Orts Escolano, Chloe Legendre, Christian Haene, Sofien
  Bouaziz, Christoph Rhemann, Paul Debevec, and Sean Fanello.
\newblock Total relighting: learning to relight portraits for background
  replacement.
\newblock {\em TOG}, 2021.

\bibitem{park2019semantic}
Taesung Park, Ming-Yu Liu, Ting-Chun Wang, and Jun-Yan Zhu.
\newblock Semantic image synthesis with spatially-adaptive normalization.
\newblock In {\em CVPR}, pages 2337--2346, 2019.

\bibitem{peng2021animatable}
Sida Peng, Junting Dong, Qianqian Wang, Shangzhan Zhang, Qing Shuai, Xiaowei
  Zhou, and Hujun Bao.
\newblock Animatable neural radiance fields for modeling dynamic human bodies.
\newblock In {\em ICCV}, 2021.

\bibitem{peng2021neural}
Sida Peng, Yuanqing Zhang, Yinghao Xu, Qianqian Wang, Qing Shuai, Hujun Bao,
  and Xiaowei Zhou.
\newblock {N}eural {B}ody: {I}mplicit neural representations with structured
  latent codes for novel view synthesis of dynamic humans.
\newblock In {\em CVPR}, 2021.

\bibitem{raj2020anr}
Amit Raj, Julian Tanke, James Hays, Minh Vo, Carsten Stoll, and Christoph
  Lassner.
\newblock {ANR}-articulated neural rendering for virtual avatars.
\newblock In {\em CVPR}, 2021.

\bibitem{ramamoorthi2001sh}
Ravi Ramamoorthi and Pat Hanrahan.
\newblock An efficient representation for irradiance environment maps.
\newblock In {\em SIGGRAPH}, 2001.

\bibitem{saito2020pifuhd}
Shunsuke Saito, Tomas Simon, Jason Saragih, and Hanbyul Joo.
\newblock Pifuhd: Multi-level pixel-aligned implicit function for
  high-resolution 3d human digitization.
\newblock In {\em CVPR}, 2020.

\bibitem{NeuralFace2017}
Z. Shu, E. Yumer, S. Hadap, K. Sunkavalli, E. Shechtman, and D. Samaras.
\newblock Neural face editing with intrinsic image disentangling.
\newblock In {\em CVPR}, 2017.

\bibitem{shysheya2019textured}
Aliaksandra Shysheya, Egor Zakharov, Kara-Ali Aliev, Renat Bashirov, Egor
  Burkov, Karim Iskakov, Aleksei Ivakhnenko, Yury Malkov, Igor Pasechnik,
  Dmitry Ulyanov, et~al.
\newblock Textured {N}eural {A}vatars.
\newblock In {\em CVPR}, 2019.

\bibitem{starck2007surface}
Jonathan Starck and Adrian Hilton.
\newblock Surface capture for performance-based animation.
\newblock {\em IEEE {C}omputer {G}raphics and {A}pplications}, 2007.

\bibitem{su2021anerf}
Shih-Yang Su, Frank Yu, Michael Zollh{\"o}fer, and Helge Rhodin.
\newblock {A-NeRF}: {A}rticulated neural radiance fields for learning human
  shape, appearance, and pose.
\newblock In {\em NeurIPS}, 2021.

\bibitem{sun2019single}
Tiancheng Sun, Jonathan~T Barron, Yun-Ta Tsai, Zexiang Xu, Xueming Yu, Graham
  Fyffe, Christoph Rhemann, Jay Busch, Paul~E Debevec, and Ravi Ramamoorthi.
\newblock Single image portrait relighting.
\newblock In {\em SIGGRAPH}, 2019.

\bibitem{tajimaPG21}
Daichi Tajima, Yoshihiro Kanamori, and Yuki Endo.
\newblock Relighting humans in the wild: Monocular full-body human relighting
  with domain adaptation.
\newblock {\em Computer Graphics Forum}, 2021.

\bibitem{tewari2022advances}
Ayush Tewari, Justus Thies, Ben Mildenhall, Pratul Srinivasan, Edgar Tretschk,
  W Yifan, Christoph Lassner, Vincent Sitzmann, Ricardo Martin-Brualla, Stephen
  Lombardi, et~al.
\newblock Advances in neural rendering.
\newblock {\em Computer Graphics Forum}, 41(2):703--735, 2022.

\bibitem{wang2022arah}
Shaofei Wang, Katja Schwarz, Andreas Geiger, and Siyu Tang.
\newblock {ARAH}: Animatable volume rendering of articulated human sdfs.
\newblock In {\em ECCV}, 2022.

\bibitem{wang2021dance}
Tuanfeng~Y Wang, Duygu Ceylan, Krishna~Kumar Singh, and Niloy~J Mitra.
\newblock Dance in the wild: Monocular human animation with neural dynamic
  appearance synthesis.
\newblock In {\em 3DV}, 2021.

\bibitem{wang2004ssim}
Z. {Wang}, A.~C. {Bovik}, H.~R. {Sheikh}, and E.~P. {Simoncelli}.
\newblock Image quality assessment: From error visibility to structural
  similarity.
\newblock {\em TIP}, 2004.

\bibitem{wang2020single}
Zhibo Wang, Xin Yu, Ming Lu, Quan Wang, Chen Qian, and Feng Xu.
\newblock Single image portrait relighting via explicit multiple reflectance
  channel modeling.
\newblock In {\em ToG}, 2020.

\bibitem{weng2022humannerf}
Chung-Yi Weng, Brian Curless, Pratul~P Srinivasan, Jonathan~T Barron, and Ira
  Kemelmacher-Shlizerman.
\newblock Human{N}erf: Free-viewpoint rendering of moving people from monocular
  video.
\newblock In {\em CVPR}, 2022.

\bibitem{yang2021towards}
Lingbo Yang, Pan Wang, Chang Liu, Zhanning Gao, Peiran Ren, Xinfeng Zhang,
  Shanshe Wang, Siwei Ma, Xiansheng Hua, and Wen Gao.
\newblock Towards fine-grained human pose transfer with detail replenishing
  network.
\newblock In {\em TIP}, 2021.

\bibitem{yang2021s3}
Ze Yang, Shenlong Wang, Sivabalan Manivasagam, Zeng Huang, Wei-Chiu Ma, Xinchen
  Yan, Ersin Yumer, and Raquel Urtasun.
\newblock S3: Neural shape, skeleton, and skinning fields for 3d human
  modeling.
\newblock In {\em CVPR}, 2021.

\bibitem{yeh2022learning}
Yu-Ying Yeh, Koki Nagano, Sameh Khamis, Jan Kautz, Ming-Yu Liu, and Ting-Chun
  Wang.
\newblock Learning to relight portrait images via a virtual light stage and
  synthetic-to-real adaptation.
\newblock {\em SIGGRAPH ASIA}, 2022.

\bibitem{yoon2022motion}
Jae~Shin Yoon, Duygu Ceylan, Tuanfeng~Y. Wang, Jingwan Lu, Jimei Yang, Zhixin
  Shu, and Hyun~Soo Park.
\newblock Learning motion-dependent appearance for high-fidelity rendering of
  dynamic humans from a single camera.
\newblock In {\em CVPR}, 2022.

\bibitem{zhang2018mixup}
Hongyi Zhang, Moustapha Cisse, Yann~N. Dauphin, and David Lopez-Paz.
\newblock mixup: Beyond empirical risk minimization.
\newblock In {\em ICLR}, 2018.

\bibitem{zhang2018perceptual}
Richard Zhang, Phillip Isola, Alexei~A Efros, Eli Shechtman, and Oliver Wang.
\newblock The unreasonable effectiveness of deep features as a perceptual
  metric.
\newblock In {\em CVPR}, 2018.

\bibitem{zhang2021neural}
Xiuming Zhang, Sean Fanello, Yun-Ta Tsai, Tiancheng Sun, Tianfan Xue, Rohit
  Pandey, Sergio Orts-Escolano, Philip Davidson, Christoph Rhemann, Paul
  Debevec, et~al.
\newblock Neural light transport for relighting and view synthesis.
\newblock {\em ACM Transactions on Graphics (TOG)}, 2021.

\bibitem{zhao2022high}
Hao Zhao, Jinsong Zhang, Yu-Kun Lai, Zerong Zheng, Yingdi Xie, Yebin Liu, and
  Kun Li.
\newblock High-fidelity human avatars from a single rgb camera.
\newblock In {\em CVPR}, 2022.

\bibitem{zhi2020texmesh}
Tiancheng Zhi, Christoph Lassner, Tony Tung, Carsten Stoll, Srinivasa~G
  Narasimhan, and Minh Vo.
\newblock Texmesh: Reconstructing detailed human texture and geometry from
  rgb-d video.
\newblock In {\em ECCV}, 2020.

\bibitem{zhou2019dpr}
Hao Zhou, Sunil Hadap, Kalyan Sunkavalli, and David~W. Jacobs.
\newblock Deep single portrait image relighting.
\newblock In {\em ICCV}, 2019.

\end{thebibliography}
